\documentclass[final,5p,times,twocolumn]{elsarticle}

\usepackage{xparse}
\usepackage{amsmath}

\usepackage{amssymb}
\usepackage{amsfonts}
\usepackage{pgfplots}
\usepackage{bbm}
\usepackage{color}
\usepackage{tikz}
\usepackage{natbib}

\usetikzlibrary{plotmarks}

\usepackage{graphicx}
\graphicspath{{./}}
\usepackage{caption}
\usepackage[colorlinks,bookmarksopen,bookmarksnumbered,citecolor=red,urlcolor=red]{hyperref}
\usepackage[capitalize]{cleveref}
\usepackage[caption=false]{subfig}

\usepackage{url}
\usepackage{framed}

\usepackage{algpseudocode}
\usepackage{algorithm}
\usepackage{amsthm}

\usepackage{threeparttable}
\usepackage{booktabs}
\usepackage{arydshln}

\usepackage{balance}
\usepackage{enumerate}

\usepackage{pdfpages}

\newcommand{\bLambda}{\boldsymbol{\Lambda}}
\newcommand{\bSigma}{\boldsymbol{\Sigma}}

\newcommand{\bPi}{\boldsymbol{\varpi}}

\newtheorem{definition}{Definition}
\newtheorem{lemma}{Lemma}

\newtheorem{proposition}{Proposition}

\journal{Robotics and Autonomous Systems}

\begin{document}

\begin{frontmatter}

\title{A Riemannian Metric for Geometry-Aware\\ Singularity Avoidance by Articulated Robots}

\author[label2,label1]{Filip Mari\'{c}\corref{correspondingauthor}}
\ead{filip.maric@fer.hr}

\author[label1]{Luka Petrovi\'{c}}
\ead{luka.petrovic@fer.hr}

\author[label1]{Marko Guberina}
\ead{marko.guberina@fer.hr}

\author[label2]{Jonathan Kelly}
\ead{jonathan.kelly@robotics.utias.utoronto.ca}

\author[label1,label2]{Ivan Petrovi\'{c}}
\ead{ivan.petrovic@fer.hr}

\address[label2]{University of Toronto Institute for Aerospace Studies,\\
Space and Terrestrial Autonomous Robotic Systems (STARS) Laboratory,\\
4925 Dufferin Street, Toronto, Ontario M4Y 1G3, Canada\\}

\address[label1]{University of Zagreb Faculty of Electrical Engineering and Computing,\\
Laboratory for Autonomous Systems and Mobile Robotics (LAMOR),\\
Unska 3, HR-10000, Zagreb, Croatia,\\}

\cortext[correspondingauthor]{Corresponding author: Filip Mari\'{c}}

\nonumnote{This research has been supported by the European Regional Development Fund under the grant KK.01.1.1.01.0009 (DATACROSS) and by the Canada Research Chairs program.}

\begin{abstract}
Articulated robots such as manipulators increasingly must operate in uncertain and dynamic environments where interaction (with human coworkers, for example) is necessary.
In these situations, the capacity to quickly adapt to unexpected changes in operational space constraints is essential.
At certain points in a manipulator's configuration space, termed \textit{singularities}, the robot loses one or more degrees of freedom (DoF) and is unable to move in specific operational space directions.
The inability to move in arbitrary directions in operational space compromises adaptivity and, potentially, safety.
We introduce a \textit{geometry-aware singularity index}, defined using a Riemannian metric on the manifold of symmetric positive definite matrices, to provide a measure of proximity to singular configurations.
We demonstrate that our index avoids some of the failure modes and difficulties inherent to other common indices.
Further, we show that our index can be differentiated easily, making it compatible with local optimization approaches used for operational space control.
Our experimental results establish that, for reaching and path following tasks, optimization based on our index outperforms a common manipulability maximization technique and ensures singularity-robust motions.
\end{abstract}


\end{frontmatter}

\section{Introduction}
Articulated robots are often required to perform tasks in which operational space movement is constrained, due to safety considerations or for other reasons.
The constraints may also be altered during task execution as a result of unexpected changes in the environment (such as a human coworker pushing the robot, for example).
Depending on the link and joint geometry, certain joint configurations can lead to a loss of operational space mobility or to hazardous joint movements, potentially resulting in task failure.
Such configurations are known as \textit{singularities}~\cite{duffy1980analysis} and singularity avoidance is an important part of most control and motion planning algorithms for articulated robots.
Identifying and avoiding singularities has thus been the focus of significant research efforts within the robotics community~\cite{tourassis1992identification}.
Moreover, geometrically intuitive optimization criteria that encode the proximity of a configuration to one or more singular or near-singular regions have found a variety of applications, ranging from control and motion planning to kinematic synthesis.

A majority of articulated robots are comprised of revolute joints, and nearly all relevant tasks can be represented by sets of nonlinear constraints that are a function of the joint positions.
When these constraints are defined in operational space, robots are especially vulnerable to kinematic singularities~\cite{beiner1992singularity, beiner1997singularity, buss2004introduction}.
Kinematic singularities inhibit the robot's ability to generate end-effector velocities in certain directions in operational space.
They can be identified by observing the conditioning of the Jacobian matrix of the robot, which maps configuration space velocities to operational space velocities~\cite{sciavicco2012modelling}.
This concept forms the basis of several kinematic sensitivity indices proposed in literature~\cite{cardou2010kinematic, patel2015manipulator}.
Many such indices can be interpreted geometrically through the notion of the \textit{manipulability ellipsoid}~\cite{yoshikawa1985manipulability}, whose axis lengths correspond to the singular values of the Jacobian matrix and indicate the  overall sensitivity to actuator displacements.
Perhaps the most common index is the \textit{manipulability index}~\cite{yoshikawa1985manipulability} proposed by Yoshikawa, which is proportional to the volume of the manipulability ellipsoid~\cite{maciejewski1989singular}.
Salisbury and Craig suggest a \textit{dexterity index} in \cite{salisbury1982articulated} that provides an upper bound for the relative error amplification, which is a function of the ratio between the longest and shortest manipulability ellipsoid axes.
A geometry-aware similarity measure between two manipulability ellipsoids is formulated by Rozo et al.\ in \cite{rozo2017learning} using the Stein divergence.

In this paper, we introduce a \textit{geometry-aware singularity index} based on a differential geometric characterization of the manipulability ellipsoid described by Jaquier et al.\ \cite{jaquier2020geometry}, which can be made robust to the failure modes of the manipulability and dexterity indices. 
We base our index on the Riemannian metric in~\cite{pennec2006riemannian}, enabling us to compute the length of the geodesic between the manipulability ellipsoid and a sufficiently ``non-singular'' reference ellipsoid.
By determining the gradient of this length with respect to the joint values, we are able to augment common operational space tracking methods with an effective singularity avoidance criterion.

Most operational space tracking schemes for robotic manipulators are feedback-based formulations that use a linearized kinematic model---this approach has been successfully applied to control, planning, and inverse kinematics~\cite{sciavicco2012modelling}.
Such methods use the local mapping of joint motions to spatial displacements (i.e., the Jacobian) of the robot to produce the desired end-effector movement~\cite{xian2004task, pham2010position}.
For inverse kinematics, local joint displacements calculated in this manner can be applied iteratively until convergence, with the end-effector pose error serving as the feedback signal~\cite{buss2004introduction, beeson2015trac}.
Null-space optimization techniques~\cite{nakamura1987task} have proven to be very effective when resolving kinematic redundancies by using the extra DoFs for optimizing secondary objectives.
These redundancy resolution schemes~\cite{nakamura1987task, marani2002real} have long been relied upon for singularity avoidance, however they are subject to algorithmic singularities \cite{chiaverini1997singularity} caused by contradictory objectives and hence are commonly used only for simple task hierarchies.
Methods based on quadratic programming (QP)~\cite{frank1956algorithm} can easily be extended to include a variety of additional constraints and objectives~\cite{cheng1993obstacle, schulman2013finding}.
Recently, these formulations have been explored as an efficient method for singularity avoidance in constrained inverse kinematics solvers~\cite{zhang2012manipulability, dufour2017integrating, jin2017manipulability}.
We integrate our geometry-aware singularity index within such a QP formulation and show through experimental evaluation that our proposed approach outperforms a similar manipulability maximization method~\cite{dufour2017integrating} for reaching and path following tasks using a variety of manipulators.
In summary, the main contributions of this paper are as follows:
\begin{enumerate}
  \item We introduce a \textit{singularity index} based on the length of the geodesic between the current manipulability ellipsoid and a reference ellipsoid.
  \item Using a result from computational matrix analysis, we show that this index can be easily differentiated with respect to joint values and used for singularity avoidance.
  \item We analyze the geometric interpretation of our index for two distinct choices of reference ellipsoid and demonstrate that, through these choices, we avoid typical failures associated with common, related indices.
\end{enumerate}
The remainder of the paper is organized as follows.
We begin by presenting the requisite background material on kinematics and kinematic singularities in \cref{sec:background}.
In \cref{sec:singularity_index} we describe the manifold of SPD matrices and derive our geometry-aware singularity index. We then define two index parameterizations that have clear geometric interpretations and that can be used to avoid singularities in operational space tracking applications.
Next, in \cref{sec:indices}, we show how our proposed index is integrated in a QP-based operational space tracking formulation and explain the gradient computation process.
Finally, in \cref{sec:res}, we review discuss our experimental results.
\begin{figure}
  \centering
  \includegraphics[page=1,width=0.7\columnwidth]{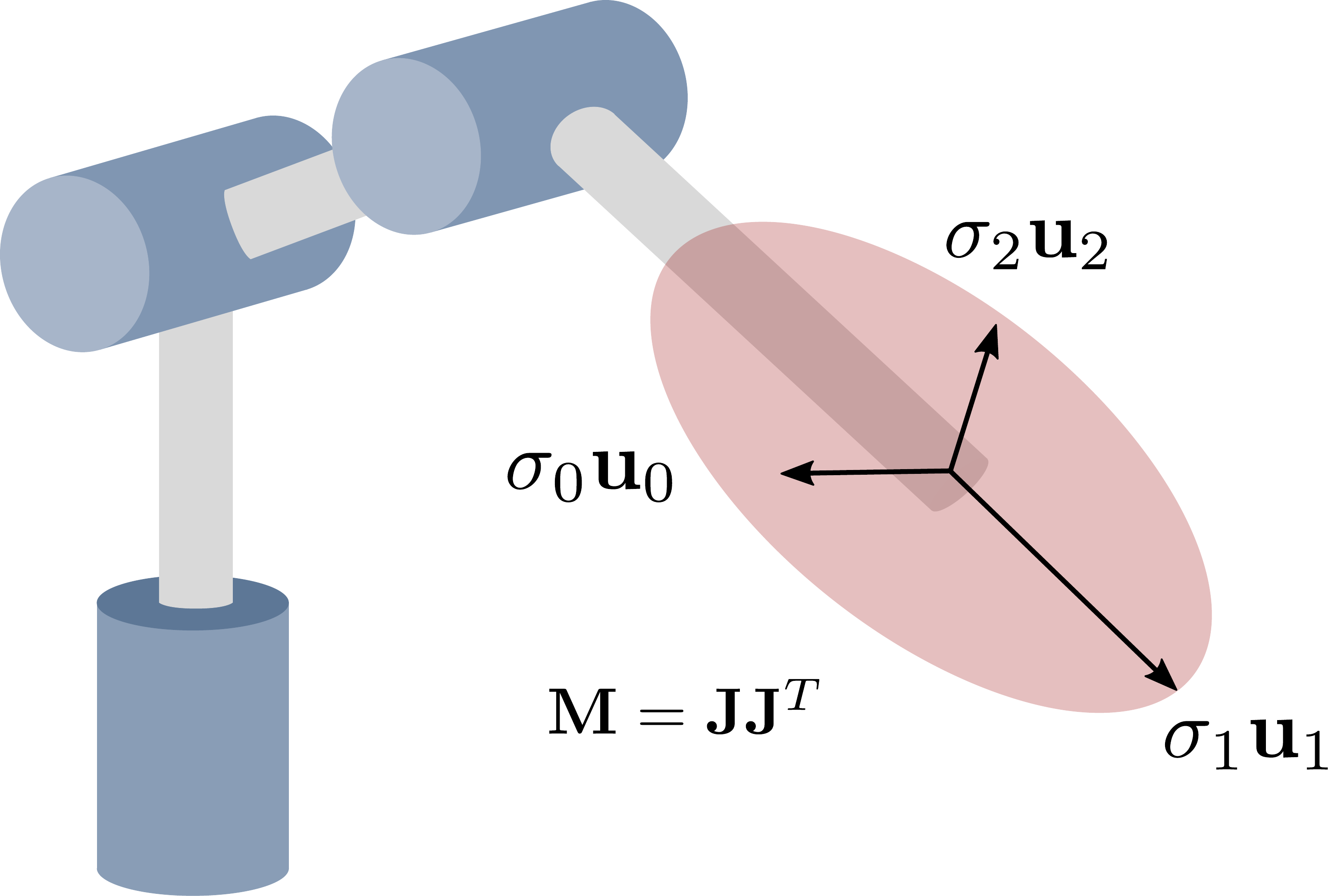}
  \caption{A three DoF manipulator and the manipulability ellipsoid associated with an operational space defined by the end-effector position. Note that the axes of the ellipsoid correspond to the singular values $\sigma$ and vectors $\mathbf{u}$ of the Jacobian $\mathbf{J}$. If we were to include the end-effector orientation, the ellipsoid would be six-dimensional.}
\label{fig:mnpel}
\vspace{-2mm}
\end{figure}

\section{Background}\label{sec:background}
Consider a serial robotic manipulator comprised of $n$ actuated joints, which we use as a model mechanism in the remainder of the paper.
We begin by defining the \textit{configuration} of a manipulator as any set of joint positions $\mathbf{q} \in \mathcal{Q}$, where the configuration space $\mathcal{Q}$ is the space of all feasible joint positions.
Similarly, the operational space $\mathcal{X}$ is the space of poses $\mathbf{x}$ of manipulator end-effector(s) or other links (specific to a given task).
The generally nonlinear and non-convex mapping
\begin{equation*}
\mathbf{f}: \mathcal{Q} \rightarrow \mathcal{X}
\end{equation*}
defines the \textit{forward kinematics} of the manipulator.
Conversely, the inverse mapping
\begin{equation*}
\mathbf{f}^{-1}: \mathcal{X} \rightarrow \mathcal{Q}
\end{equation*}
defines the \textit{inverse kinematics} of the manipulator.
Taking the gradient of $\mathbf{f}$ with respect to the joint values $\mathbf{q}$, we arrive at the manipulator Jacobian matrix
\begin{equation*}
  \mathbf{J} = \frac{\partial\,{\mathbf{f}}}{\partial\,{\mathbf{q}}} \in \mathbb{R}^{p \times n}\, .
\end{equation*}
The kinematic relationship between configuration space and operational space velocities is defined by the linear equation
\begin{equation}\label{eq:jacobian}
  \dot{\mathbf{x}} = \mathbf{J}_{0} \dot{\mathbf{q}},
\end{equation}
where $\mathbf{J}_{0}$ is the manipulator Jacobian matrix computed at a particular configuration $\mathbf{q}_{0}$, while $\dot{\mathbf{q}} \in \mathbb{R}^{n}$ and $\dot{\textbf{x}} \in \mathbb{R}^{p}$ are the joint and operational space velocities  (at some time $t$), respectively \cite{sciavicco2012modelling}.
In the remainder of the paper, we omit explicit references to linearization points and assume that $\mathbf{J}$ represents the Jacobian computed at the current joint configuration, where applicable.

The problem of operational space tracking involves computing the joint motions required to move the end-effector in some direction in the operational space.
When the operational space reference being tracked is a desired end-effector velocity $\dot{\mathbf{x}}$, this computation can be stated in the form of a QP
\begin{equation}\label{eq:qp1}
\begin{aligned}
& \underset{\dot{\mathbf{q}}}{\text{min}}
& & \frac{1}{2}\dot{\mathbf{q}}^T\mathbf{W}\dot{\mathbf{q}} + \mathbf{w}\dot{\mathbf{q}}\\
& ~\text{s.t.} & & \mathbf{J}\dot{\mathbf{q}} = \dot{\mathbf{x}} \;\\
& & & \dot{\mathbf{q}}_{min} \leq \dot{\mathbf{q}} \leq \dot{\mathbf{q}}_{max} \; .\\
\end{aligned}
\end{equation}
The cost function in \cref{eq:qp1} serves to minimize joint velocities with respect to the symmetric weighting matrix $\mathbf{W}$, and the linear term $\mathbf{w}$ is commonly used to encode secondary objectives.
Joint velocity is limited by the linear inequality constraints $\dot{\mathbf{q}}_{min}$ and $\dot{\mathbf{q}}_{max}$.
Finally, the linear equality constraint enforces a desired end-effector velocity.
Taking $\mathbf{w} = \mathbf{0}$ and dropping the joint velocity constraints, elementary matrix calculus leads to the closed-form solution
\begin{equation}\label{eq:jacinv}
  \dot{\mathbf{q}} = \mathbf{W} \mathbf{J}^{-1} \dot{\mathbf{x}}\, ,
\end{equation}
that can be applied to control the end-effector movement.
Moreover, this formulation can be used sequentially to solve the inverse kinematics problem~\cite{schulman2013finding}.
However, the Jacobian inverse $\mathbf{J}^{-1}$ is generally numerically unstable and may result in prohibitively large and incorrect joint velocities.
Most approaches instead resort taking the pseudoinverse or manually modifying the lowest singular values of the Jacobian matrix, sacrificing tracking accuracy for numerical stability~\cite{buss2004introduction}.
In \cref{sec:mnpel}, we show how the numerical stability of the manipulator Jacobian matrix can be characterized and geometrically interpreted, allowing us to develop a more principled method for achieving robust tracking.

\subsection{The Manipulability Ellipsoid}\label{sec:mnpel}
Consider an $n$-dimensional unit sphere in the space of joint velocities $\|\dot{\mathbf{q}}\|^2 = 1$.
Observing~\cref{eq:jacinv} and assuming without loss of generality that $\mathbf{W} = \mathbf{I}$, we obtain a mapping of this sphere to the operational velocity space
\begin{equation}\label{eq:ellipsoid}
 \dot{\mathbf{q}}^{T} \dot{\mathbf{q}} = \dot{\mathbf{x}}^T\left(\mathbf{J}\mathbf{J}^T\right)^{-1}\dot{\mathbf{x}}\, .
\end{equation}
From Eq.\ \eqref{eq:ellipsoid}, it is obvious that the scaling of operational space velocities to the joint space is determined by the matrix
\begin{equation}\label{eq:mnp_el_def}
  \mathbf{M}\left(\mathbf{q}\right) = \mathbf{J}\mathbf{J}^T\, ,
\end{equation}
whose eigenvalues correspond to the squared singular values $\sigma^2$ of the Jacobian matrix $\mathbf{J}$.
Consequently, configurations in which one or more eigenvalues of $\mathbf{M}$ become  zero correspond to cases where the Jacobian matrix is badly conditioned and non-invertible.
That is, for any configuration $\mathbf{q}$, we can use Eq.\ \eqref{eq:mnp_el_def} to compute a symmetric positive semidefinite matrix $\mathbf{M}$ that contains information about the operational space mobility of the manipulator.

Notably, there exists an isomorphism between the set of $p \times p$ symmetric positive semidefinite matrices and the set of centered ellipsoids of dimension $\leq p$.
For this reason, the matrix $\mathbf{M}$ is also known as the \textit{manipulability ellipsoid} of the end-effector \cite{yoshikawa1985manipulability}.
The principal axes $\sigma_0\mathbf{u}_0, \sigma_1 \mathbf{u}_1,\dots, \sigma_p\mathbf{u}_p$ of this ellipsoid can be determined through singular value decomposition of $\mathbf{J} = \mathbf{U}\boldsymbol{\Sigma}\mathbf{V}^T$.
The lengths and orientations of these axes indicate directions in which greater operational space velocities can be generated.
This is illustrated in \cref{fig:mnpel}, where the operational space consists of the end-effector position of a three DoF manipulator.
Conversely, directions admitting higher mobility are also directions in which the manipulator is more sensitive to perturbations.

\subsection{Singularity Indices}\label{subsec:singularities}
\begin{figure}
\centering
  \subfloat[]{  \includegraphics[width=0.48\columnwidth]{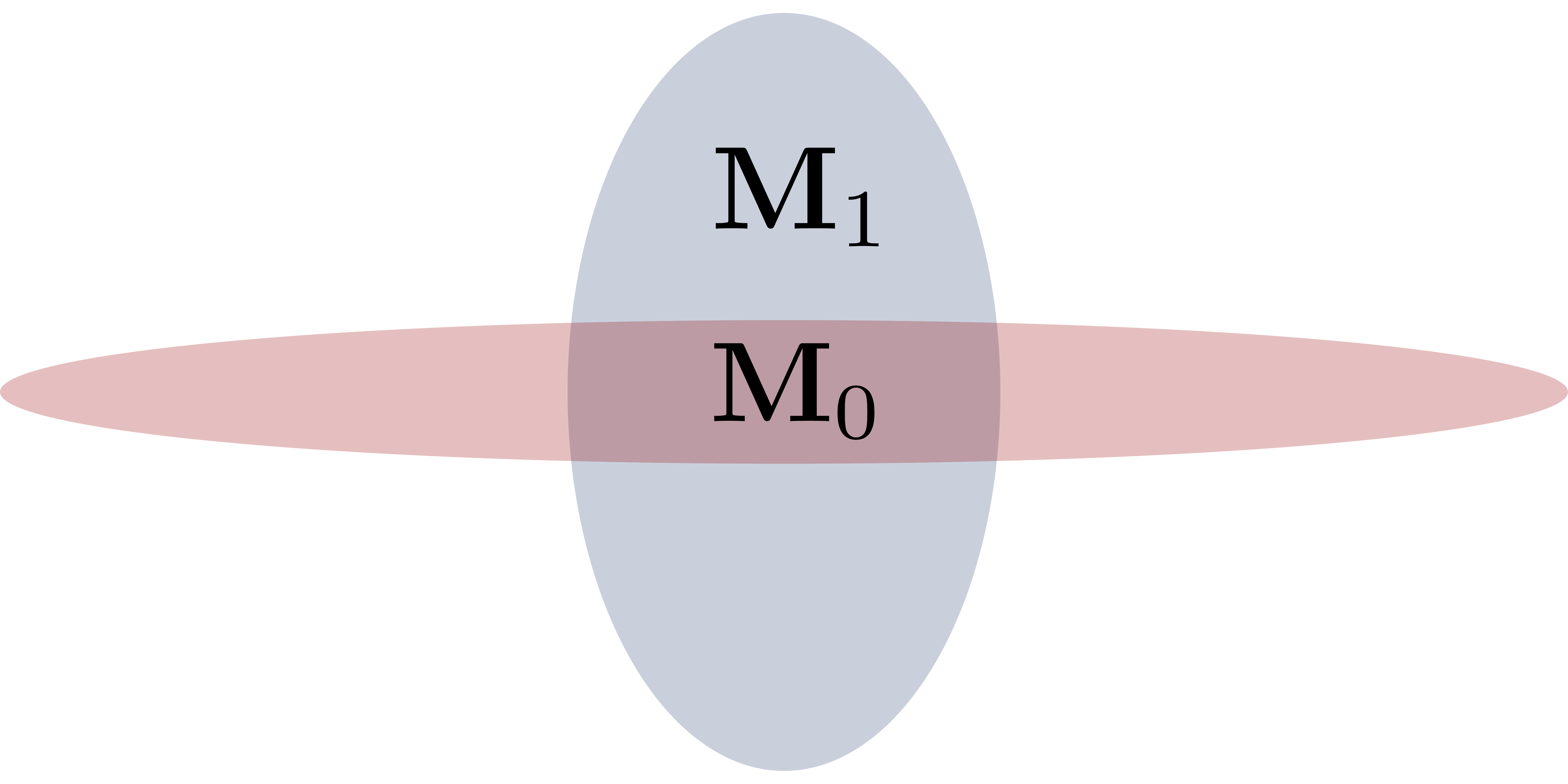}
\label{fig:degen_mnp}}
  \subfloat[]{  \includegraphics[width=0.48\columnwidth]{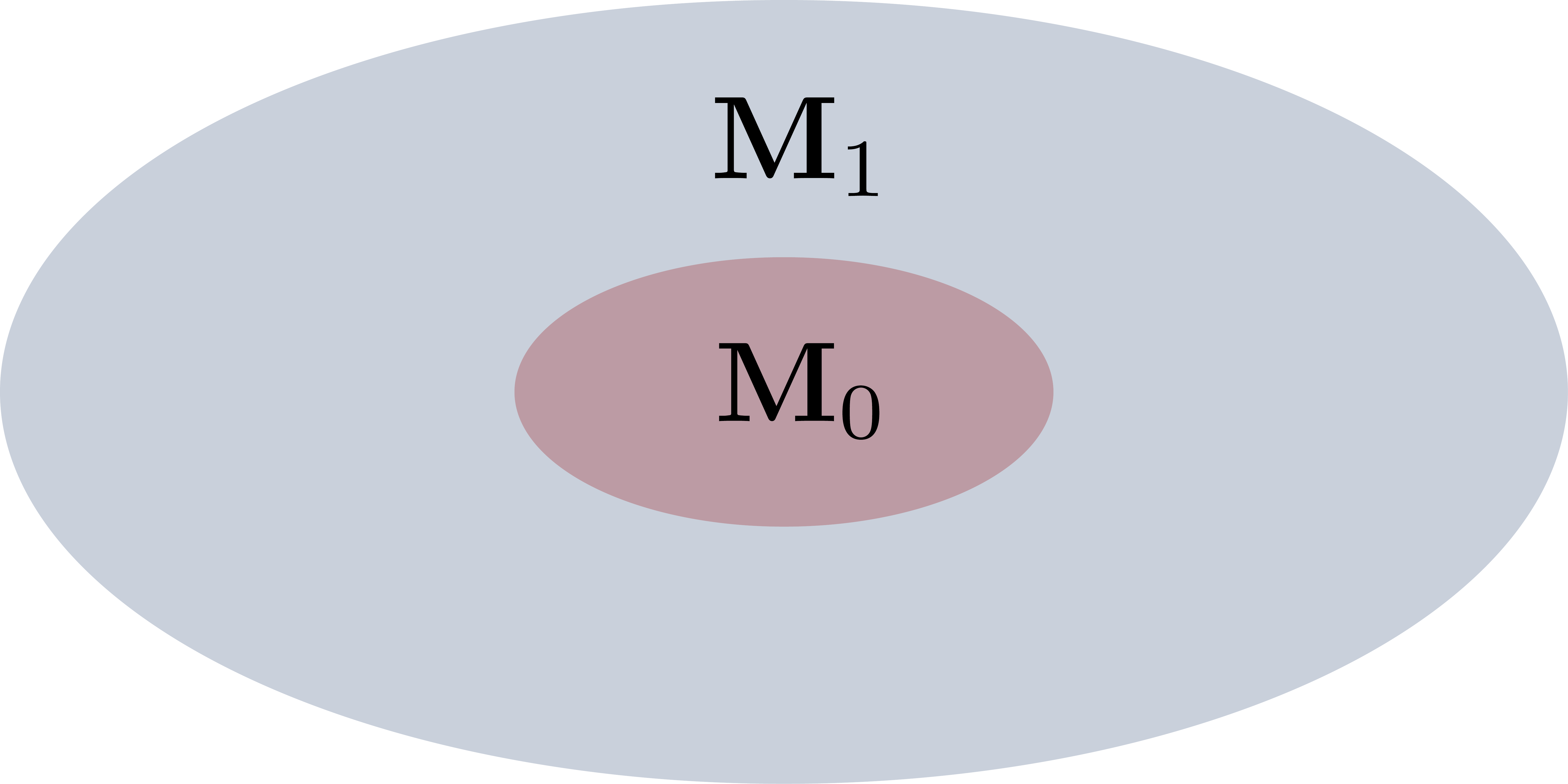}
\label{fig:degen_dex}}
\caption{An example of failure modes for common singularity indices. (a) The manipulability index fails for configurations involving elongated ellipsoids with near zero-length axes. (b) The dexterity index fails when the ellipsoids are of uniformly small scale.}
\vspace{-2mm}
\end{figure}
Following the geometric interpretation described above, it can be concluded that all configurations for which the Jacobian matrix cannot be inverted have an associated manipulability ellipsoid with one or more degenerate (i.e., zero-length or near zero-length) axes, rendering \cref{eq:jacinv} numerically unstable.
The detection and avoidance of these configurations, known as \textit{singularities}, requires careful interpretation of the Jacobian's singular values.
Moreover, end-effector movements that begin from configurations in close proximity to singularities also tend to result in high joint velocities and undesirable dynamic characteristics~\cite{cardou2010kinematic,patel2015manipulator}.
Consequently, many indicators have been developed that are used for singularity avoidance and to improve the kinematic sensitivity of operational space control and inverse kinematics algorithms.

A common indicator used to detect the proximity of a configuration to a singularity is known as the \textit{manipulability index}, expressed as
\begin{equation}\label{eq:manipulability}
m = \sqrt{\det(\mathbf{J} \mathbf{J}^{T})}\, .
\end{equation}
The manipulability index also admits a geometric interpretation, since the index value is proportional to the volume of the manipulability ellipsoid.
Because singularities involve manipulability ellipsoids with one or more axes of zero length, and therefore zero volume, the manipulability index can be used to detect such configurations.
The differentiability of \cref{eq:manipulability} has resulted in many approaches for singularity avoidance that optimize the manipulability index in order to maximize the volume of the manipulability ellipsoid~\cite{dufour2017integrating,maric2016robot}.
Manipulability is not as effective in detecting configurations in close proximity to singularities due to scenarios such as the one shown in \cref{fig:degen_mnp}, where the volume of the ellipsoid remains relatively large despite the lengths of certain axes being close to zero.

Another common index used to detect and avoid singularities is known as \textit{dexterity index}~\cite{salisbury1982articulated},
\begin{equation}
\kappa = \frac{\sigma_{max}}{\sigma_{min}}\, ,
\end{equation}
where $\sigma_{max}$ and $\sigma_{min}$ are the maximal and minimal singular values of the Jacobian.
It follows that the dexterity index value is a measure of distortion of manipulator sensitivity in Cartesian space in a given configuration.
That is, the index is the measure of the difference in relative lengths between the longest and shortest axes of the manipulability ellipsoid. 
The geometric interpretation again reveals that an important drawback of this approach lies in the inability to encode the scale of the ellipsoid.
Since the ratio of axis lengths becomes infinite only when the configuration is exactly singular and gives no information about the ellipsoid size, it is impossible to use the dexterity index to provide a measure of proximity to a singularity, as show in \cref{fig:degen_dex}.
Note that there exist many other measures of kinematic sensitivity used in singularity avoidance and tailored for specific problem instances such as parallel manipulators or walking robots~\cite{cardou2010kinematic}.

\section{A Geometry-Aware Singularity Index}
\label{sec:singularity_index}

The manipulability and dexterity indices described in Section~\ref{subsec:singularities} are commonly used for detecting and avoiding singular configurations.
In this section we show that the differential-geometric characterization of manipulability ellipsoids introduced in ~\cite{jaquier2020geometry} induces a Riemannian metric that naturally defines a distance between manipulability ellipsoids.
We use this distance to specify a family of geometrically intuitive singularity indices, parameterized by the choice of reference ellipsoid.

\subsection{The Riemannian Manifold of SPD Matrices}
\label{subsec:spd}
\begin{figure}
  \centering
  \includegraphics[page=1,width=0.5\columnwidth]{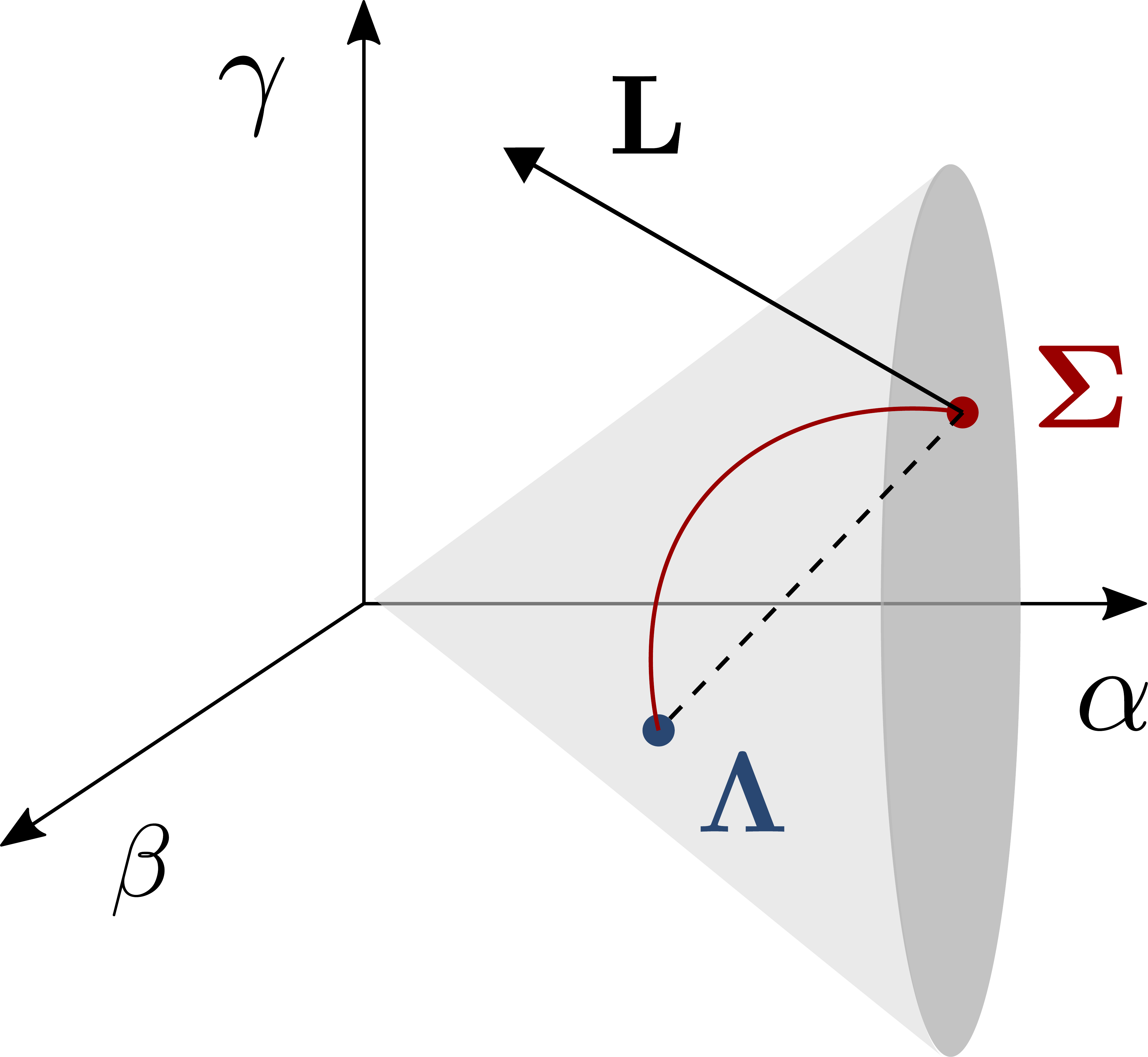}%
  \caption{Visualization of the convex cone formed by the set $\mathcal{S}_{++}^2$ of matrices of the form $\protect\begin{pmatrix} \alpha & \beta \\ \beta & \gamma \protect\end{pmatrix}$. The matrices $\bSigma$ and $\bLambda$ lie inside the cone, and the matrix $\mathbf{L} = \mbox{Log}_{\bSigma}(\bLambda)$ lies in the tangent space of $\bSigma$. The shortest path connecting $\bSigma$ and $\bLambda$ is the geodesic shown in red. Note that the length of the geodesic differs from the length of the dashed straight line in Euclidean space.}
  \label{fig:PSD}
  \vspace{-2mm}
\end{figure}
Manipulability ellipsoids of non-singular configurations correspond to the set of symmetric matrices with strictly positive eigenvalues.
This set is known as the set of symmetric positive definite (SPD) matrices,
\begin{equation}
\mathcal{S}_{++}^p = \{\bSigma \, |\, \bSigma = \bSigma^{T},\, \mathbf{x}^{T}\bSigma \mathbf{x} > 0 \, \forall \mathbf{x} \in \mathbb{R}^{p}\}\, ,
\end{equation}
which forms a convex cone in $\mathbb{R}^{K}$, where $K = p(p+1)/2$.
As shown in \cref{fig:PSD}, a straight line in $\mathbb{R}^{K}$ does not represent the shortest path between points on the $\mathcal{S}_{++}^p$ manifold.
This means that we cannot rely on the Euclidean metric to induce a distance that is useful when reasoning about the similarity of ellipsoids.
Fortunately, the field of Riemannian geometry equips us with the tools necessary to establish an alternative, Riemannian metric on the set $\mathcal{S}_{++}^p$, forming a Riemannian manifold.
In turn, we are able to define a geometrically-appropriate distance on this manifold.

\begin{definition}[Riemannian manifold \cite{lee2018introduction}]

A Riemannian manifold $\mathcal{M}$ is a smooth manifold equipped with a positive-definite inner product $\langle \mathbf{Z}_1,\mathbf{Z}_2 \rangle_{\bSigma} $ on the tangent space $T_{\bSigma}\mathcal{M}$ of each point $\bSigma \in \mathcal{M}$ that varies smoothly from point to point.

\end{definition}
As shown in \cref{fig:PSD}, the tangent space of elements in $\mathcal{M} \equiv \mathcal{S}_{++}^p$ is the space of symmetric matrices $T_{\bSigma}\mathcal{M} \equiv \operatorname{Sym}_p$, which is a vector space in $\mathbb{R}^K$.
The positive-definite inner product $\langle \mathbf{Z}_1,\mathbf{Z}_2 \rangle_{\bSigma}$ defined on this tangent space is also known as the Riemannian metric.
\begin{definition}[Riemannian metric on $\mathcal{S}_{++}^p$ \cite{pennec2006riemannian}]
  For some $\bSigma \in \mathcal{M}$, a positive-definite inner product of two elements $\mathbf{Z}_{1},\mathbf{Z}_{2} \in T_{\bSigma}\mathcal{M}$ can be defined as
  \begin{equation}\label{eq:riemmanian_metric}
    \langle \mathbf{Z}_{1}, \mathbf{Z}_{2} \rangle_{\bSigma} = \operatorname{Tr}(\bSigma^{-\frac{1}{2}}\mathbf{Z}_{1}\,\bSigma^{-1}\,\mathbf{Z}_{2}\, \bSigma^{-\frac{1}{2}})\, ,
  \end{equation}
  and is called the Riemannian metric on $\mathcal{M}$.
\end{definition}
\noindent An important property of this metric is its invariance to affine transformations \cite{pennec2006riemannian}:
for any $\mathbf{A} \in GL_{p}$ with the group action $\mathbf{A}\star \bSigma = \mathbf{A}\bSigma\mathbf{A}^{T}$ on the space $\mbox{Sym}_p$, we have
\begin{equation}\label{eq:invariance}
\langle \mathbf{A}\star \mathbf{Z}_{1}, \mathbf{A} \star \mathbf{Z}_{2}\rangle_{\mathbf{A}\star \bSigma} = \langle \mathbf{Z}_{1}, \mathbf{Z}_{2} \rangle_{\bSigma}.
\end{equation}
This property will be useful when considering how the manipulability ellipsoid varies depending on configuration.

The Riemannian metric allows us to determine the lengths of curves on the manifold.
When considering the similarity of ellipsoids, we are particularly interested in geodesic curves.
\begin{definition}[Geodesic~\cite{lee2018introduction}]
  Consider a parametric curve $\gamma(t)$ on $\mathcal{M}$ and its velocity vector $\dot{\gamma}(t)$ that lies on $T_{\gamma(t)}\mathcal{M}$.
  The length of this curve over $t \in [0, T]$ is computed as:
  \begin{equation*}
    \mathcal{L}(\gamma) = \int_{0}^{T} \left(\langle \dot{\gamma}(t), \dot{\gamma}(t)\rangle_{\gamma(t)}\right)^{\frac{1}{2}}
  \end{equation*}
  The shortest path between two elements $\bSigma, \bLambda \in \mathcal{M}$ corresponds to the locally length-minimizing curve
  \begin{equation}\label{eq:geodesic}
    d(\bSigma, \bLambda) = \min_{\gamma(0) = \bSigma,\, \gamma(T) = \bLambda} \mathcal{L}(\gamma)\, ,
  \end{equation}
  called the \textit{geodesic}.
\end{definition}
\noindent The length of the geodesic connecting two elements of a Riemannian manifold is known as the Riemannian distance.
\begin{definition}[Riemannian distance on $\mathcal{S}_{++}^p$ \cite{pennec2006riemannian}]
  For $\mathcal{M}$ equipped with the Riemannian metric of \cref{eq:riemmanian_metric}, the Riemannian distance between $\bSigma, \bLambda \in \mathcal{M}$ is defined as
  \begin{equation} \label{eq:af_dist}
    d(\bSigma, \bLambda) =  \left\lVert\log\left(\bSigma^{-\frac{1}{2}}\bLambda\bSigma^{-\frac{1}{2}} \right)\right\rVert_F\, .
  \end{equation}
\end{definition}
In fact, this is the length of the geodesic connecting two non-singular manipulability ellipsoids~\cite{jaquier2020geometry}.

\subsection{Riemannian Distance as a Singularity Index}\label{subsec:riem_dist_as_sing}
In \cref{subsec:spd} we showed that the Riemannian distance between the manipulability ellipsoid $\mathbf{M}$ and a reference ellipsoid $\bSigma$ can be obtained using \cref{eq:af_dist}.
If we assume a geometrically appropriate $\bSigma$ is chosen, we can apply \cref{eq:af_dist} to obtain a notion of proximity of a given configuration to a singularity.
Consider the manipulability ellipsoid $\mathbf{M}\left(\mathbf{q}\right) = \mathbf{J}\mathbf{J}^T$ of a robot in some configuration $\mathbf{q} \in \mathcal{Q}$ and some reference ellipsoid $\bSigma$.
We define the squared length of the geodesic (i.e., the squared Riemannian distance) connecting $\mathbf{M}$ and $\bSigma$ as
\begin{equation}\label{eq:riem_avoid}
\xi = \left\lVert\mbox{log}\left(\bSigma^{-\frac{1}{2}}\mathbf{M}\bSigma^{-\frac{1}{2}} \right)\right\rVert^2_F \, ,
\end{equation}
which follows directly from \cref{eq:af_dist}.
In the context of singularity detection and avoidance, we refer to the scalar $\xi$ in \cref{eq:riem_avoid} as the geometry-aware singularity index.

As a robot moves and changes its configuration, the manipulability ellipsoid varies in shape, size, and orientation.
In \cref{subsec:singularities} we noted that conventional indices such as \textit{dexterity} and \textit{manipulability} generally represent only one particular property of the manipulability ellipsoid (i.e., volume or axis length ratio).
In contrast, $\xi$ is a function of all axis lengths, as well as the ellipsoid's shape and orientation.
This presents an opportunity to avoid many of the issues that are common when applying the conventional indices; the problems arise because one property (e.g., the ellipsoid volume) may remain constant while the ellipsoid itself changes.
Importantly, the affine-invariance property of the underlying Riemannian metric described in \cref{subsec:spd} guarantees that the value of our index is always determined by the relative difference between two ellipsoids.
It is therefore crucial that our choice of reference ellipsoid reflects the goal of singularity avoidance.%
\subsection{Choosing the Reference Ellipsoid $\bSigma$}
The utility of the geometry-aware singularity index in \cref{eq:riem_avoid} clearly depends on an appropriate choice of the reference ellipsoid $\bSigma$, which must be selected with the goal of singularity avoidance in mind.
In this section, we propose two possible choices that help to avoid some of the degeneracies that may occur when using other common indices.%
\subsubsection{Choosing $\bSigma = k\, \mathbf{I}$}\label{sec:circle}
\begin{figure}
\centering
\includegraphics[width=0.6\columnwidth]{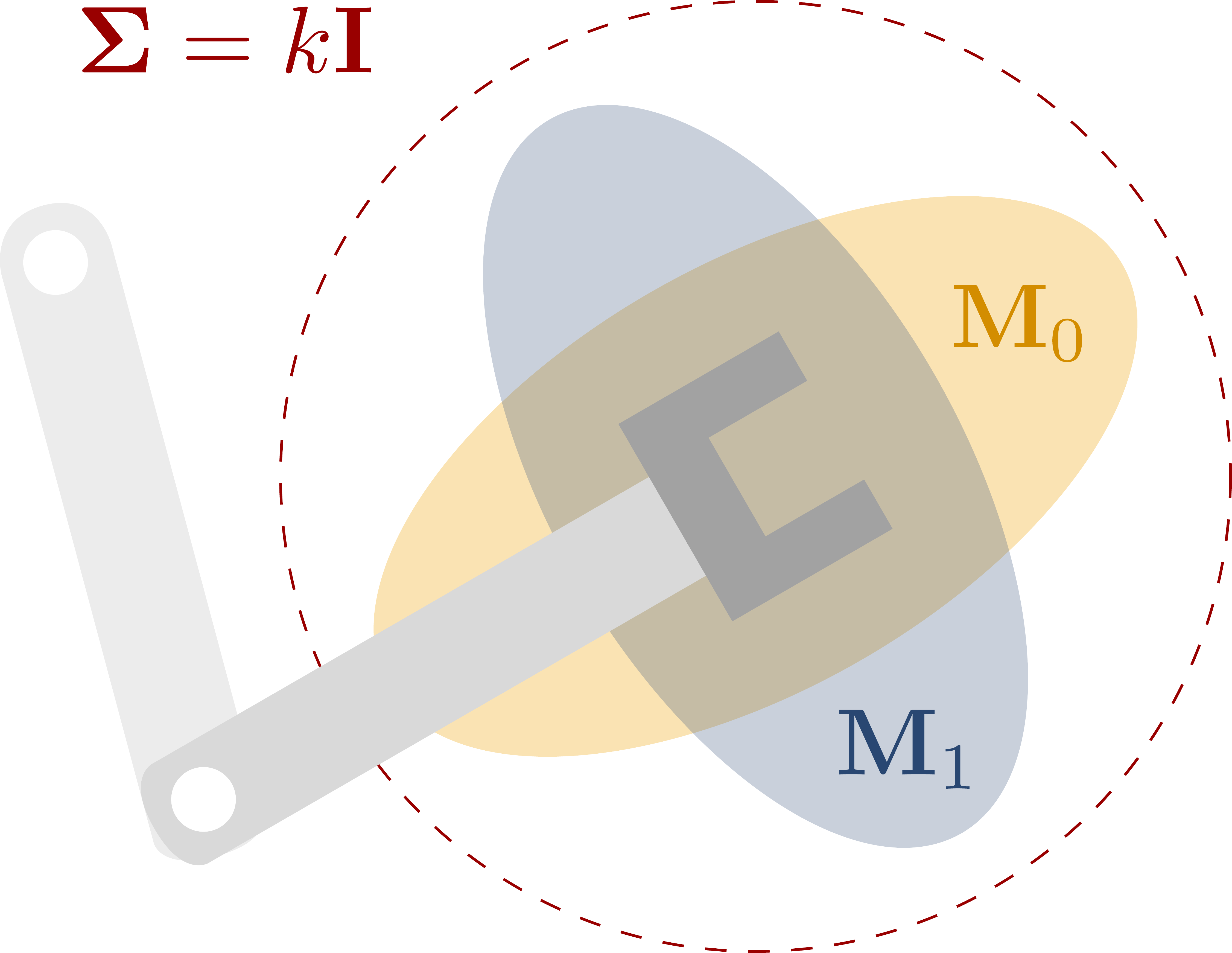}
\caption{Singularity avoidance formulation s-IK, in which the distance between a sphere $\bSigma = k\, \mathbf{I}$ and the manipulability ellipsoid $\mathbf{M}$ is minimized. Note that both $\mathbf{M}_0$ and $\mathbf{M}_1$ are the same distance from $\bSigma$, since the squared metric  $\xi$ is independent of orientation. }
\label{fig:PSD3}
\vspace*{-2mm}
\end{figure}
Consider selecting a reference ellipsoid $\bSigma$ that has a spherical shape with a radius greater or equal to the length of the longest possible manipulability ellipsoid axis, as shown in \cref{fig:PSD3}.
Formally, this class of ellipsoids can be expressed as
\begin{equation}\label{eq:s-IK}
	\bSigma = k\mathbf{I}, \, k\geq \sigma_{max}^2\, .
\end{equation}
The scaling factor $k$ in \cref{eq:s-IK} is chosen to be larger than the largest manipulability ellipsoid eigenvalue, forming a sphere that encapsulates the ellipsoid.
By choosing
  \begin{equation}\label{eq:k}
    k \geq \text{Tr}\left( \mathbf{M}(\mathbf{q}) \right)\,, \forall \mathbf{q} \in \mathcal{Q},
  \end{equation}
we ensure that the geometry-aware singularity index (denoted by $\xi$) decreases with the increase of the singular values of the Jacobian, since the manipulability ellipsoid will always be contained within the reference ellipsoid $\bSigma$.
Because the maximum volume of the manipulability ellipsoid for any manipulator is bounded, $k$ can also be found empirically by moving the robot and increasing $k$ whenever \cref{eq:k} fails to hold.

Since the sphere is symmetric, $\xi$ is invariant to the orientation of the manipulability ellipsoid.
This result follows directly from the affine-invariance property described in \cref{subsec:riem_dist_as_sing} and can be proven easily by inserting $\bSigma = k\, \mathbf{I}$ into \cref{eq:invariance}.
Such a property is desirable in a singularity avoidance context because the orientation of the manipulability ellipsoid does not change the singular values of the Jacobian.

\begin{figure}[!t]
  \captionsetup[subfloat]{position=top,labelformat=empty}
  \centering
  \subfloat[]{  \resizebox{0.4\textwidth}{!}{\input{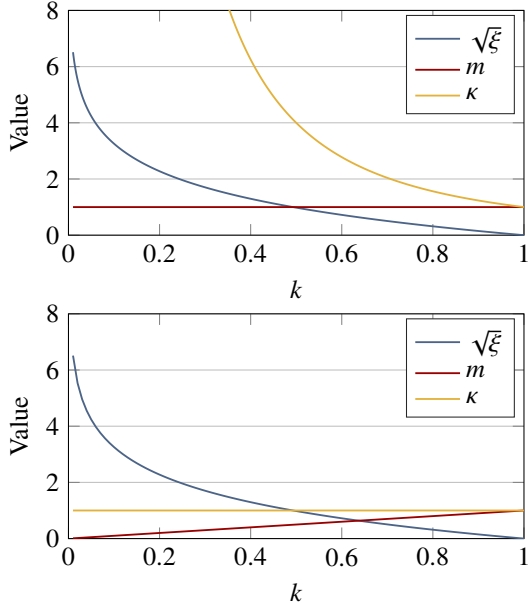}}
  } \\
  \vspace{-1cm}
  \subfloat[]{  \resizebox{0.4\textwidth}{!}{
%
%
\definecolor{blue}{RGB}{76,100,135}
\definecolor{red}{RGB}{153,0,0}
\definecolor{yellow}{RGB}{227,178,60}
\definecolor{mycolor1}{rgb}{0.00000,0.44700,0.74100}%
\definecolor{mycolor2}{rgb}{0.85000,0.32500,0.09800}%
\definecolor{mycolor3}{rgb}{0.92900,0.69400,0.12500}%
\begin{tikzpicture}

\begin{axis}[%
width=6.634in,
height=3.612in,
at={(2.596in,2.358in)},
xmin=0,
xmax=1,
xlabel style={font=\color{white!15!black}},
xlabel={$k$},
xlabel near ticks,
ymin=0,
ymax=8,
ylabel style={font=\color{white!15!black}},
ylabel={Value},
ylabel near ticks,
ymajorgrids,
scale=0.4,
axis background/.style={fill=white},
legend style={legend cell align=left, align=left, draw=white!15!black}
]
\addplot [color=blue, line width = 0.25mm]
  table[row sep=crcr]{%
1	0\\
0.99	0.0142133212704263\\
0.98	0.0285709426850902\\
0.97	0.043075824324011\\
0.96	0.057731018293665\\
0.95	0.0725396725816698\\
0.94	0.08750503511543\\
0.93	0.102630458037836\\
0.92	0.117919402214096\\
0.91	0.133375441984851\\
0.9	0.149002270181921\\
0.89	0.164803703424261\\
0.88	0.180783687713158\\
0.87	0.196946304347162\\
0.86	0.21329577617895\\
0.85	0.22983647423811\\
0.84	0.246572924745832\\
0.83	0.263509816549668\\
0.82	0.280652009008924\\
0.81	0.298004540363841\\
0.8	0.315572636624652\\
0.79	0.333361721019731\\
0.78	0.351377424045593\\
0.77	0.369625594165324\\
0.76	0.388112309206322\\
0.75	0.406843888512908\\
0.74	0.425826905914631\\
0.73	0.445068203576843\\
0.72	0.464574906806573\\
0.71	0.484354439893882\\
0.7	0.504414543076819\\
0.69	0.524763290727004\\
0.68	0.545409110862762\\
0.67	0.566360806107829\\
0.66	0.587627576226066\\
0.65	0.60921904237658\\
0.64	0.631145273249304\\
0.63	0.65341681325874\\
0.62	0.676044712993475\\
0.61	0.699040562141598\\
0.6	0.72241652513756\\
0.59	0.746185379804995\\
0.58	0.770360559302801\\
0.57	0.79495619771923\\
0.56	0.819987179701471\\
0.55	0.845469194557053\\
0.54	0.871418795319481\\
0.53	0.897853463334949\\
0.52	0.924791679001232\\
0.51	0.95225299937567\\
0.5	0.980258143468547\\
0.49	1.00882908615364\\
0.48	1.03798916176221\\
0.47	1.06776317858398\\
0.46	1.09817754568264\\
0.45	1.12926041365047\\
0.44	1.1610418311817\\
0.43	1.1935539196475\\
0.42	1.22683106821438\\
0.41	1.26091015247747\\
0.4	1.2958307800932\\
0.39	1.33163556751414\\
0.38	1.36837045267487\\
0.37	1.40608504938318\\
0.36	1.44483305027512\\
0.35	1.48467268654537\\
0.34	1.52566725433131\\
0.33	1.56788571969461\\
0.32	1.61140341671785\\
0.31	1.65630285646202\\
0.3	1.70267466860611\\
0.29	1.75061870277135\\
0.28	1.80024532317002\\
0.27	1.85167693878803\\
0.26	1.90504982246978\\
0.25	1.96051628693709\\
0.24	2.01824730523076\\
0.23	2.07843568915119\\
0.22	2.14129997465025\\
0.21	2.20708921168293\\
0.2	2.27608892356175\\
0.19	2.34862859614342\\
0.18	2.42509119374367\\
0.17	2.50592539779986\\
0.16	2.5916615601864\\
0.15	2.68293281207466\\
0.14	2.78050346663857\\
0.13	2.88530796593833\\
0.12	2.99850544869931\\
0.11	3.1215581181188\\
0.1	3.25634706703029\\
0.09	3.40534933721221\\
0.0800000000000001	3.57191970365494\\
0.0700000000000001	3.76076161010711\\
0.0599999999999999	3.97876359216786\\
0.05	4.23660521049884\\
0.04	4.55217784712349\\
0.03	4.9590217356364\\
0.02	5.53243599059204\\
0.01	6.51269413406059\\
};
\addlegendentry{$\sqrt{\xi}$}

\addplot [color=red, line width = 0.25mm]
  table[row sep=crcr]{%
1	1\\
0.99	0.99\\
0.98	0.98\\
0.97	0.97\\
0.96	0.96\\
0.95	0.95\\
0.94	0.94\\
0.93	0.93\\
0.92	0.92\\
0.91	0.91\\
0.9	0.9\\
0.89	0.89\\
0.88	0.88\\
0.87	0.87\\
0.86	0.86\\
0.85	0.85\\
0.84	0.84\\
0.83	0.83\\
0.82	0.82\\
0.81	0.81\\
0.8	0.8\\
0.79	0.79\\
0.78	0.78\\
0.77	0.77\\
0.76	0.76\\
0.75	0.75\\
0.74	0.74\\
0.73	0.73\\
0.72	0.72\\
0.71	0.71\\
0.7	0.7\\
0.69	0.69\\
0.68	0.68\\
0.67	0.67\\
0.66	0.66\\
0.65	0.65\\
0.64	0.64\\
0.63	0.63\\
0.62	0.62\\
0.61	0.61\\
0.6	0.6\\
0.59	0.59\\
0.58	0.58\\
0.57	0.57\\
0.56	0.56\\
0.55	0.55\\
0.54	0.54\\
0.53	0.53\\
0.52	0.52\\
0.51	0.51\\
0.5	0.5\\
0.49	0.49\\
0.48	0.48\\
0.47	0.47\\
0.46	0.46\\
0.45	0.45\\
0.44	0.44\\
0.43	0.43\\
0.42	0.42\\
0.41	0.41\\
0.4	0.4\\
0.39	0.39\\
0.38	0.38\\
0.37	0.37\\
0.36	0.36\\
0.35	0.35\\
0.34	0.34\\
0.33	0.33\\
0.32	0.32\\
0.31	0.31\\
0.3	0.3\\
0.29	0.29\\
0.28	0.28\\
0.27	0.27\\
0.26	0.26\\
0.25	0.25\\
0.24	0.24\\
0.23	0.23\\
0.22	0.22\\
0.21	0.21\\
0.2	0.2\\
0.19	0.19\\
0.18	0.18\\
0.17	0.17\\
0.16	0.16\\
0.15	0.15\\
0.14	0.14\\
0.13	0.13\\
0.12	0.12\\
0.11	0.11\\
0.1	0.1\\
0.09	0.09\\
0.0800000000000001	0.0800000000000001\\
0.0700000000000001	0.0700000000000001\\
0.0599999999999999	0.0599999999999999\\
0.05	0.05\\
0.04	0.04\\
0.03	0.03\\
0.02	0.02\\
0.01	0.01\\
};
\addlegendentry{$m$}

\addplot [color=yellow, line width = 0.25mm]
  table[row sep=crcr]{%
1	1\\
0.99	1\\
0.98	1\\
0.97	1\\
0.96	1\\
0.95	1\\
0.94	1\\
0.93	1\\
0.92	1\\
0.91	1\\
0.9	1\\
0.89	1\\
0.88	1\\
0.87	1\\
0.86	1\\
0.85	1\\
0.84	1\\
0.83	1\\
0.82	1\\
0.81	1\\
0.8	1\\
0.79	1\\
0.78	1\\
0.77	1\\
0.76	1\\
0.75	1\\
0.74	1\\
0.73	1\\
0.72	1\\
0.71	1\\
0.7	1\\
0.69	1\\
0.68	1\\
0.67	1\\
0.66	1\\
0.65	1\\
0.64	1\\
0.63	1\\
0.62	1\\
0.61	1\\
0.6	1\\
0.59	1\\
0.58	1\\
0.57	1\\
0.56	1\\
0.55	1\\
0.54	1\\
0.53	1\\
0.52	1\\
0.51	1\\
0.5	1\\
0.49	1\\
0.48	1\\
0.47	1\\
0.46	1\\
0.45	1\\
0.44	1\\
0.43	1\\
0.42	1\\
0.41	1\\
0.4	1\\
0.39	1\\
0.38	1\\
0.37	1\\
0.36	1\\
0.35	1\\
0.34	1\\
0.33	1\\
0.32	1\\
0.31	1\\
0.3	1\\
0.29	1\\
0.28	1\\
0.27	1\\
0.26	1\\
0.25	1\\
0.24	1\\
0.23	1\\
0.22	1\\
0.21	1\\
0.2	1\\
0.19	1\\
0.18	1\\
0.17	1\\
0.16	1\\
0.15	1\\
0.14	1\\
0.13	1\\
0.12	1\\
0.11	1\\
0.1	1\\
0.09	1\\
0.0800000000000001	1\\
0.0700000000000001	1\\
0.0599999999999999	1\\
0.05	1\\
0.04	1\\
0.03	1\\
0.02	1\\
0.01	1\\
};
\addlegendentry{$\kappa$}

\end{axis}
\end{tikzpicture}%
}
  }
  \caption{Performance of our geometry-aware singularity index $\xi$ for $\bSigma=\text{Tr}(\mathbf{M}) \, \mathbf{I}$ in cases where the manipulability index $m$ (top) and dexterity index $\kappa$ (bottom) are ambiguous. The top plot corresponds to the constant volume scenario shown in \cref{fig:degen_mnp}, while the bottom plot represents the constant shape scenario from \cref{fig:degen_dex}.}
    \label{fig:compare}
\end{figure}

We can study the behavior of the geometry-aware singularity index with this choice of $\bSigma$ in scenarios where the more common manipulability and dexterity indices fail, as shown in \cref{fig:degen_mnp} and \cref{fig:degen_dex}.
We begin with an arbitrary ellipse that has two axes of equal length; the two axes are then progressively scaled by the factors $k$ and $1/k$, respectively, effectively `squeezing' the ellipse in a way that maintains the overall area.
In the top plot of \cref{fig:compare}, we see that the manipulability index remains unchanged because the area is constant, making it impossible to differentiate between the nearly singular `squeezed' ellipse and a circle.
The dexterity index increases with $k$ and reaches a value of $1$ for the circular shape.
Our geometry-aware singularity index decreases as $k \rightarrow 1$, since the ellipse moves closer to the circular $\bSigma$ on the manifold.
Next, we again consider an arbitrary ellipse with axes of equal length that are both progressively scaled by $k$, uniformly inflating the ellipse.
The bottom of plot of \cref{fig:compare} shows that the dexterity index does not differentiate between the smaller and larger ellipses because the ratio of the maximal and minimal axis lengths remains unchanged.
The manipulability index is proportional to the area of the ellipse and thus increases as $k \rightarrow 1$, while our geometry-aware singularity index decreases as the ellipse expands towards a circle of radius of $1$.
These results demonstrate that our proposed index avoids some notable `blind spots' of the two commonly-used indices.

\subsubsection{Choosing $\bSigma = k \mathbf{M}$}\label{sec:ellipsoid}
\begin{figure}
  \centering
\includegraphics[width=0.54\columnwidth]{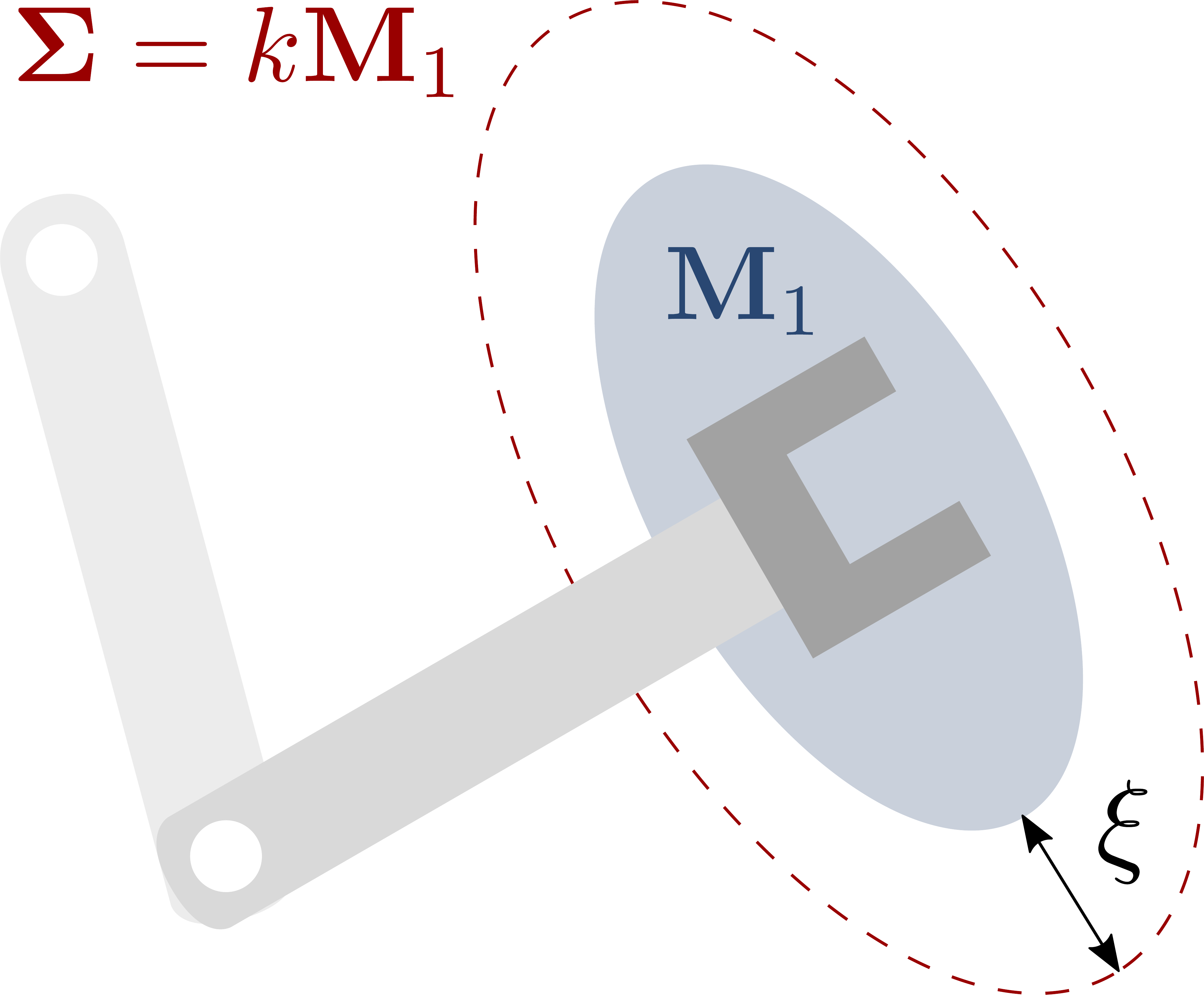}
\caption{Singularity avoidance formulation s-IK2, in which the reference ellipsoid $\bSigma$ produced at each iteration is a scaled variant of the original.}
\label{fig:PSD2}
\vspace*{-2mm}
\end{figure}
As described in \cref{sec:background}, control and planning algorithms of the form in \cref{eq:qp1} commonly integrate criteria such as singularity avoidance by directly making use of the gradient of the relevant indices.
This reveals an interesting instance of our proposed index, where the reference ellipsoid is a scaled version of the manipulability ellipsoid at the current configuration of the robot.
We begin by choosing
\begin{equation}
  \bSigma = k\mathbf{M}_{0},\,\, k\geq 1,
\end{equation}
where $\mathbf{M}_{0}$ is the manipulability ellipsoid evaluated at each time step of the tracking algorithm. Inserting $\bSigma$ into \cref{eq:riem_avoid}, the inside of the matrix logarithm evaluates to
\begin{equation}\label{eq:proof1}
\left(k\mathbf{M}_{0}\right)^{-\frac{1}{2}}\mathbf{M}\left(k\mathbf{M}_{0}\right)^{-\frac{1}{2}} \Big|_{\mathbf{M} = \mathbf{M}_{0}} = k^{-1}\mathbf{I}
\end{equation}
at each operating point.
While it is clear that our index evaluates identically at every operating point, we can gain additional insight by considering the gradient of the index.

In \cref{sec:gradient}, we explain that analytical gradient computation for our index is generally non-trivial due to the requirement that the elementwise derivative of $\bSigma^{-\frac{1}{2}}\mathbf{M}\bSigma^{-\frac{1}{2}}$ be commutative with its inverse.
However, it follows from \cref{eq:proof1} that the identity in \cref{eq:logder_no} holds at the operating point, and we are able to obtain the partial derivative
\begin{equation}\label{eq:distgrad2}
  \frac{\partial \xi}{\partial q_i} = -2~\mbox{log}(k)~\mbox{Tr}\left(\frac{\partial \mathbf{J}}{\partial q_i} \mathbf{J}^{\dagger}\right)\, ,
\end{equation}
where $\mathbf{J}^{\dagger}$ is the Jacobian pseudo-inverse.
As shown in \cref{fig:PSD2}, this gradient, constructed from partial derivatives in \cref{eq:distgrad2}, gives a direction in which $\mathbf{M}$ expands along all of its axes.
Interestingly, \cref{eq:distgrad2} is proportional to the gradient of the manipulability index \cite{maric2019fast}, revealing a differential-geometric generalization of manipulability maximization approaches commonly used in kinematic control.
In fact, our results in \cref{sec:reaching} confirm that this instance of our index performs similarly to manipulability maximization in compatible operational space tracking algorithms.
Performance for this choice of $\bSigma$ hinges on the rate at which $\mathbf{M}_{0}$ is updated, as all the above properties are lost outside the neighborhood of the operating point.

\section{Singularity Avoidance}\label{sec:indices}
In this section, we show how the geometry-aware singularity index defined by \cref{eq:af_dist} can be used for singularity avoidance in a common family of operational space control and inverse kinematics algorithms.
The formulation used herein can be seen as an extension of that in~\cite{jaquier2020geometry}, where a Jacobian-based approach is employed to follow reference directions in the tangent space of SPD matrices in order to obtain a specific, desired ellipsoid.
Reaching a specific orientation and shape of the manipulability ellipsoid is not important in general for ensuring singularity avoidance, whereas maintaining sufficient Jacobian conditioning is key.
Therefore, we directly optimize the squared affine-invariant distance between the current manipulability ellipsoid $\mathbf{M}$ and a reference ellipsoid $\bSigma$, reflected in the geometry-aware singularity index $\xi$.
Our index can be added to the cost function of a common nonlinear programming formulation of the inverse kinematics problem
\begin{equation}\label{eq:nlp}
\begin{aligned}
& \underset{\mathbf{q}}{\text{min}}
& & (\mathbf{q} - \mathbf{q}_{0})^{T}\mathbf{W}(\mathbf{q} - \mathbf{q}_{0}) + \alpha\, \xi(\mathbf{q})\\
& ~\text{s.t.}
& & \mathbf{f}(\mathbf{q})= \mathbf{T_{goal}} \in \textit{SE}(3) \; ,\\
\end{aligned}
\end{equation}
where $\mathbf{W}$ is a weighting matrix used to prioritize certain joints, $\mathbf{T}_{goal}$ is the goal end-effector pose, and $\alpha$ is a gain parameter.
A solution to \cref{eq:nlp} can be found by iteratively solving a sequence of quadratic programs obtained by linearizing the cost and constraints.
This sequential quadratic programming (SQP) approach was previously shown to be effective when designing singularity-robust kinematic controllers~\cite{dufour2017integrating,zhang2016qp}.
By adding a velocity-minimizing term to the cost and joint velocity constraints, we arrive at the following QP
\begin{equation}\label{eq:qp2}
\begin{aligned}
& \underset{\dot{\mathbf{q}}}{\text{min}}
& & \dot{\mathbf{q}}^T\mathbf{W}\dot{\mathbf{q}} + \alpha (\nabla \xi(\mathbf{q}_{0}))\,\dot{\mathbf{q}}\\
& ~\text{s.t.}
& & \quad \quad \mathbf{J}\dot{\mathbf{q}} = \dot{\mathbf{x}} \;\\
& & & \dot{\mathbf{q}}_{min} \leq \dot{\mathbf{q}} \leq \dot{\mathbf{q}}_{max} \; ,\\
\end{aligned}
\end{equation}
which is exactly the operational space tracking formulation shown in \cref{eq:qp1}.
In both control and inverse kinematics applications, this QP is redefined at each time instance or iteration and new $\nabla \xi$ and $\mathbf{J}$ are calculated to reflect the current configuration.
The joint velocity limits serve the additional purpose of enforcing joint position limits, as the velocity limits can be changed at each iteration to reflect the space of locally feasible joint motions.
Depending on the choice of the gain parameter $\alpha$, the robot will be guided in a direction in which the overall distance from singularities increases or decreases.
Note that problems of the form in \cref{eq:qp2} have been shown to allow for a wide variety of additional constraints, such as collision avoidance~\cite{schulman2013finding} and manipulability maximization~\cite{dufour2017integrating}.

\subsection{Gradient Computation}\label{sec:gradient}
Optimization methods used in control and kinematic synthesis require the gradient $\nabla \xi$ to produce joint displacements that avoid singularities.
The gradient can be expressed as a concatenation of partial derivatives of $\xi$ with respect to the joint positions $q_i$
\begin{equation*}
  \nabla \xi= \left[ \frac{\partial \xi}{\partial q_{0}} \dots \frac{\partial \xi}{\partial q_{n}} \right] \in \mathbb{R}^{n} \, .
\end{equation*}%
These partial derivatives can be obtained using elementary matrix calculus as
\begin{equation}\label{eq:partial_d}
  \frac{\partial \xi}{\partial q_i}  = 2\,\mbox{Tr}\left(\frac{\partial\log{\left(\bPi\right)} }{\partial q_i}\log{(\bPi)}^T\right)\, ,
\end{equation}
where
\begin{equation}\label{eq:pi}
  \bPi = \bSigma^{-\frac{1}{2}}\mathbf{M}\bSigma^{-\frac{1}{2}}\,.
\end{equation}
The partial derivative of $\log{(\bPi)}$ admits the closed form solution
\begin{equation}\label{eq:logder_no}
\frac{\partial\log{\left(\bPi\right)} }{\partial q_i} = \bPi^{-1}\frac{\partial \bPi}{\partial q_{i}}
\end{equation}
only if the matrices $\bPi^{-1}$ and $\frac{\partial \bPi}{\partial q_{i}}$ commute.
Unfortunately, the matrices in~\cref{eq:logder_no} are generally not commutative and the identity is therefore invalid when considering $\bSigma$ of an arbitrary shape.
Instead of finding the partial derivatives analytically, we can evaluate them numerically by leveraging a result from computational matrix analysis.%
We begin with a lemma showing that \cref{eq:logder_no} can be expressed using directional derivatives:
\begin{lemma}[\cite{dattorro2005convex}]\label{lemma:one}
  The partial derivatives of $\log{(\bPi)}$ with respect to individual elements of $\mathbf{q}$ can be defined as
  \begin{equation}\label{eq:lem1}
  \frac{\partial \log{(\bPi)}}{\partial q_{i}} = L_{ \log{(\bPi)}}\left(\bPi, \frac{\partial \bPi}{\partial q_{i}}\right)\, ,
  \end{equation}
  where $L_{\mathbf{f}}(\mathbf{g}, \mathbf{E})$ is the directional derivative of $\mathbf{f}$ at $\mathbf{g}$ in the direction $\mathbf{E}$.

  \begin{proof}
    Using the chain rule for matrix-valued functions, the partial derivative of $\mathbf{f}(\mathbf{g}(\mathbf{x})): \mathbb{R}^{n}\rightarrow \mathbb{R}^{K \times K}$ with respect to $\mathbf{x} \in \mathbb{R}^{n}$ is expressed as
    \begin{equation}\label{eq:proof1-1}
      \frac{\partial \mathbf{f}}{\partial x_{i}} = \nabla_{\mathbf{g}}\mathbf{f} \cdot \frac{\partial \mathbf{g}}{\partial x_{i}}\, .
    \end{equation}
    The product definition of the directional derivative of $\mathbf{f}$ at $\mathbf{g}$ in the $\mathbf{E}$ direction is given by
    \begin{equation}\label{eq:proof1-2}
      L_{\mathbf{f}}(\mathbf{g}, \mathbf{E}) = \nabla_{\mathbf{g}}\mathbf{f} \cdot \mathbf{E}\, .
    \end{equation}
    The equivalence of \cref{eq:proof1-1} and \cref{eq:proof1-2} is self-evident.
  \end{proof}
\end{lemma}
\noindent From \cref{lemma:one}, it follows that the partial derivative of the logarithm in \cref{eq:partial_d} can be computed using \cref{eq:lem1}.
We first take the derivative along the direction $\frac{\partial \bPi}{\partial q_{i}}$, which has a closed form expression
\begin{equation}\label{eq:bpider}
\frac{\partial \bPi}{\partial q_{i}} = \bSigma^{-\frac{1}{2}}\left(\frac{\partial \mathbf{J}}{\partial q_{i}} \mathbf{J}^{T} + \mathbf{J}\frac{\partial \mathbf{J}}{\partial q_{i}}^{T} \right)\bSigma^{-\frac{1}{2}}\, .
\end{equation}
Once the direction is obtained, the directional derivative can be accurately and efficiently computed using the identity introduced in~\cite{higham2008functions}, which we formalize in the following proposition:
\begin{proposition}
  The partial derivatives of $\log{(\bPi)}$ with respect to individual elements of $\mathbf{q}$ can be computed using the identity
  \begin{equation}\label{eq:biglog}
    \log
    \left(\,
    \begin{bmatrix}
      \bPi & \frac{\partial \bPi}{\partial q_{i}} \\
      \mathbf{0} & \bPi
    \end{bmatrix}
    \,\right)
  =
    \begin{bmatrix}
      \log{(\bPi)} & \frac{\partial \log{(\bPi)}}{\partial q_{i}}\\
      \mathbf{0} & \log{(\bPi)}
    \end{bmatrix} .
  \end{equation}
  \begin{proof}
    (Theorem 3.6 in~\cite{higham2008functions}) Let $\mathbf{f}$ be a differentiable matrix function and $\mathbf{X}(t) \in \mathbb{R}^{n \times n}$ be a symmetric matrix differentiable at $t=0$.
    Let
    \begin{equation*}
    \mathbf{X}(t) = \mathbf{X} + t\mathbf{E}\, ,
    \end{equation*}
    and we have the identity
    \begin{equation*}
      \mathbf{f}
      \left(
        \,
      \begin{bmatrix}
        \mathbf{X} & \mathbf{E} \\
        \mathbf{0} & \mathbf{X}
      \end{bmatrix}
      \,
      \right)
    =
      \begin{bmatrix}
        \mathbf{f}(\mathbf{X}) & L_{\mathbf{f}}(\mathbf{X}, \mathbf{E})\\
        \mathbf{0} & \mathbf{f}(\mathbf{X})
      \end{bmatrix} \,,
    \end{equation*}
    where $L_{\mathbf{f}}(\mathbf{X}, \mathbf{E})$ is the directional derivative of $\mathbf{f}$ at $\mathbf{X}$ in direction $\mathbf{E}$.
    From \cref{eq:pi} and \cref{eq:bpider} it is clear that for $\mathbf{X} = \bPi$ and $\mathbf{E} = \frac{\partial \bPi}{\partial q_{i}}$ the symmetry assumptions hold, completing the proof.
  \end{proof}
\end{proposition}
\noindent Since the size of the matrix representing the manipulability ellipsoid is generally at most $6 \times 6$, computing the matrix in \cref{eq:biglog} remains computationally tractable and the overall computation time negligible. %
This result allows us to explore arbitrary choices of the reference ellipsoid $\bSigma$ in \cref{eq:pi}, enabling the adaptation of $\xi$ to both the structure of the robot and the task.
\subsection{Limitations}
There are several limitations that should be considered when using the approach presented herein for singularity avoidance.
First, it is crucial that the reference ellipsoid $\bSigma$ encapsulates all manipulability ellipsoids that can be reached by the robot.
This can be accomplished by defining a large ellipsoid beforehand, either empirically or analytically, as described for the spherical reference ellipsoid in \cref{sec:circle}.
In \cref{sec:ellipsoid} we have shown that the reference ellipsoid can also be re-defined at each iteration and its lengths modified to ensure it remains larger than the manipulability ellipsoid.
This option may provide greater numerical stability, assuming that the $\bSigma$ is updated at an adequate rate.
Another problematic scenario may occur when the initial manipulator configuration is itself singular, since the manifold geometry described in \cref{sec:singularity_index} holds only for non-singular ellipsoids.
This situation is easily detected either by directly observing the singular values of the Jacobian or noting that matrix logarithm has failed---a small perturbation can be applied to the joint configuration to leave the singular region.

\section{Experimental Results}\label{sec:res}
In this section we present experimental results for the proposed geometry-aware singularity avoidance index $\xi$ when implemented within the QP-based operational space control formulation defined by \cref{eq:qp2}.
Specifically, we consider the two index variants with reference ellipsoids $\bSigma$ defined in \cref{sec:circle} (labeled s-IK) and \cref{sec:ellipsoid} (labeled s-IK2).
In order to validate the benefits of using a Riemannian metric, we also evaluate a singularity index derived from the use of the standard Euclidean metric
\begin{equation}\label{eq:euclidean}
  \xi_{E} = \lVert \mathbf{M} - \bSigma\rVert_{F}^{2}\, ,
\end{equation}
where a spherical $\bSigma$ is selected as described in \cref{sec:circle}.
As for our proposed approach, the gradient of this index is integrated into the cost \cref{eq:qp1} in place of the linear term; the resulting formulation is labeled e-IK.
We also compare our index to the manipulability maximization method from~\cite{dufour2017integrating}, where a manipulability gradient term is again added as the linear component of the cost in \cref{eq:qp1}; the resulting formulation is labeled m-IK.
As a baseline, we use a standard approach to operational space tracking obtained by setting $\alpha=0$ in \cref{eq:qp2}; this last formulation is labelled IK.

In our evaluation, we perform two benchmark experiments involving a pair of common tasks: reaching and path following.
First, in \cref{sec:reaching} we demonstrate how our method performs in a simple reaching task, where a goal end-effector position must be attained while maximizing the overall distance from singular regions.
Next, in \cref{sec:circ_path} we demonstrate how our method can be used to guide the manipulator away from singularities while following a circular end-effector path.
All experiments were performed on a laptop computer with an Intel i7-8750H CPU running at 2.20 GHz and with 16 GB of RAM.

\subsection{Reaching Task}\label{sec:reaching}
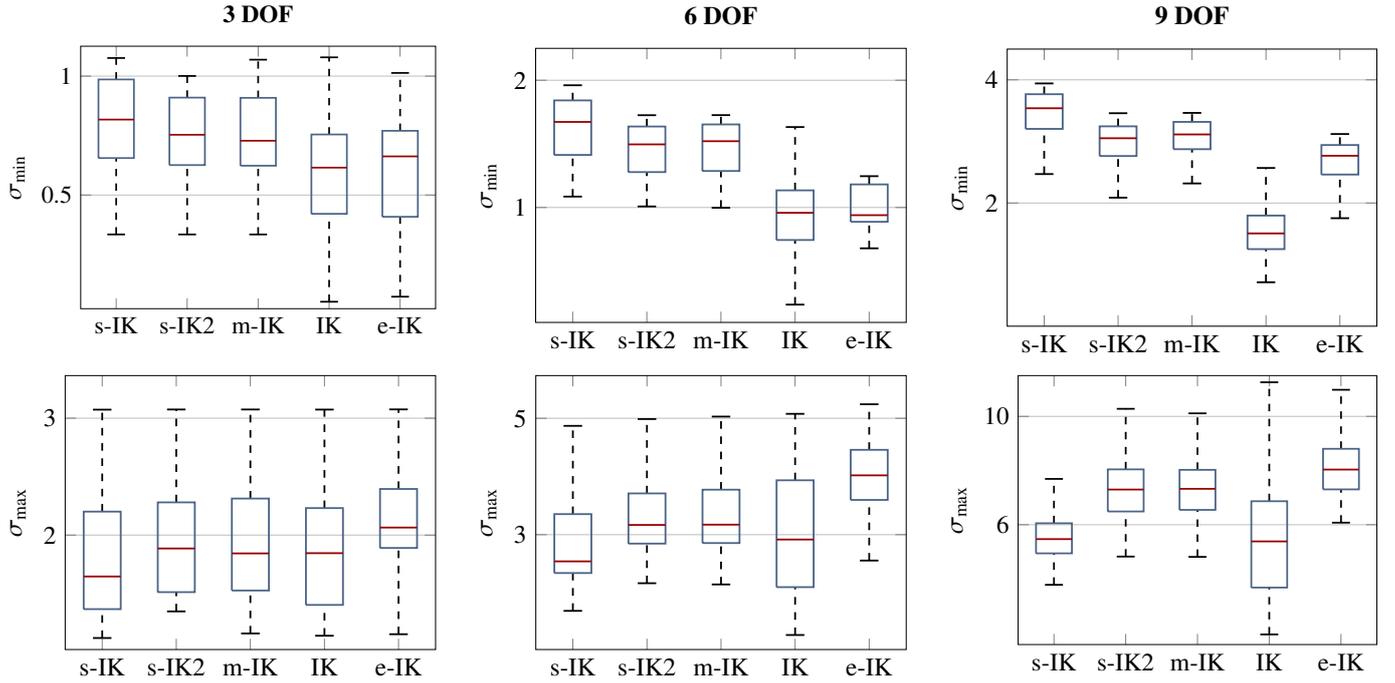
\begin{figure*}[!t]
  \captionsetup[subfloat]{position=top,labelformat=empty}
  \centering
  \hspace*{-0.6cm} 
    \resizebox{\textwidth}{!}{
  \begin{tabular}{@{}ccc@{}}
  \subfloat[]{
    \resizebox{0.33\textwidth}{!}{
%
%
\begin{tikzpicture}
\definecolor{blue}{RGB}{76,100,135}
\definecolor{red}{RGB}{153,0,0}
\begin{axis}[%
width=4.634in,
height=3.612in,
at={(0.777in,0.487in)},
unbounded coords=jump,
xmin=0.5,
xmax=5.5,
xtick={1,2,3,4,5},
xticklabels={s-IK,s-IK2,m-IK,IK, e-IK},
ymin=0.022042434369705,
ymax=1.12568981576325,
ytick={ 0.5, 1},
yticklabels={0.5, 1},
ylabel style={font=\color{white!15!black}},
ylabel={$\sigma{}_{\text{min}}$},
ylabel near ticks,
ymajorgrids,
scale=0.5,
axis background/.style={fill=white},
title style={font=\bfseries},
title={3 DOF},
legend style={legend cell align=left, align=left, draw=white!15!black}
]
\addplot [color=black, dashed, line width=0.25mm, forget plot]
  table[row sep=crcr]{%
1	0.98596269394996\\
1	1.07606792756861\\
};
\addplot [color=black, dashed, line width=0.25mm, forget plot]
  table[row sep=crcr]{%
2	0.90960086910595\\
2	1.00061908716277\\
};
\addplot [color=black, dashed, line width=0.25mm, forget plot]
  table[row sep=crcr]{%
3	0.908654807860282\\
3	1.06904984480471\\
};
\addplot [color=black, dashed, line width=0.25mm, forget plot]
  table[row sep=crcr]{%
4	0.75464756565356\\
4	1.07897288653232\\
};
\addplot [color=black, dashed, line width=0.25mm, forget plot]
  table[row sep=crcr]{%
5	0.769962145581353\\
5	1.01328551779642\\
};
\addplot [color=black, dashed, line width=0.25mm, forget plot]
  table[row sep=crcr]{%
1	0.334249197733925\\
1	0.655413078020615\\
};
\addplot [color=black, dashed, line width=0.25mm, forget plot]
  table[row sep=crcr]{%
2	0.334199187980872\\
2	0.625882472553624\\
};
\addplot [color=black, dashed, line width=0.25mm, forget plot]
  table[row sep=crcr]{%
3	0.334161027798865\\
3	0.623282656153892\\
};
\addplot [color=black, dashed, line width=0.25mm, forget plot]
  table[row sep=crcr]{%
4	0.0512853429523047\\
4	0.420977663341156\\
};
\addplot [color=black, dashed, line width=0.25mm, forget plot]
  table[row sep=crcr]{%
5	0.072434241280531\\
5	0.408502457853301\\
};
\addplot [color=black, line width=0.25mm, forget plot]
  table[row sep=crcr]{%
0.875	1.07606792756861\\
1.125	1.07606792756861\\
};
\addplot [color=black, line width=0.25mm, forget plot]
  table[row sep=crcr]{%
1.875	1.00061908716277\\
2.125	1.00061908716277\\
};
\addplot [color=black, line width=0.25mm, forget plot]
  table[row sep=crcr]{%
2.875	1.06904984480471\\
3.125	1.06904984480471\\
};
\addplot [color=black, line width=0.25mm, forget plot]
  table[row sep=crcr]{%
3.875	1.07897288653232\\
4.125	1.07897288653232\\
};
\addplot [color=black, line width=0.25mm, forget plot]
  table[row sep=crcr]{%
4.875	1.01328551779642\\
5.125	1.01328551779642\\
};
\addplot [color=black, line width=0.25mm, forget plot]
  table[row sep=crcr]{%
0.875	0.334249197733925\\
1.125	0.334249197733925\\
};
\addplot [color=black, line width=0.25mm, forget plot]
  table[row sep=crcr]{%
1.875	0.334199187980872\\
2.125	0.334199187980872\\
};
\addplot [color=black, line width=0.25mm, forget plot]
  table[row sep=crcr]{%
2.875	0.334161027798865\\
3.125	0.334161027798865\\
};
\addplot [color=black, line width=0.25mm, forget plot]
  table[row sep=crcr]{%
3.875	0.0512853429523047\\
4.125	0.0512853429523047\\
};
\addplot [color=black, line width=0.25mm, forget plot]
  table[row sep=crcr]{%
4.875	0.072434241280531\\
5.125	0.072434241280531\\
};
\addplot [color=blue, line width=0.25mm, forget plot]
  table[row sep=crcr]{%
0.75	0.655413078020615\\
0.75	0.98596269394996\\
1.25	0.98596269394996\\
1.25	0.655413078020615\\
0.75	0.655413078020615\\
};
\addplot [color=blue, line width=0.25mm, forget plot]
  table[row sep=crcr]{%
1.75	0.625882472553624\\
1.75	0.90960086910595\\
2.25	0.90960086910595\\
2.25	0.625882472553624\\
1.75	0.625882472553624\\
};
\addplot [color=blue, line width=0.25mm, forget plot]
  table[row sep=crcr]{%
2.75	0.623282656153892\\
2.75	0.908654807860282\\
3.25	0.908654807860282\\
3.25	0.623282656153892\\
2.75	0.623282656153892\\
};
\addplot [color=blue, line width=0.25mm, forget plot]
  table[row sep=crcr]{%
3.75	0.420977663341156\\
3.75	0.75464756565356\\
4.25	0.75464756565356\\
4.25	0.420977663341156\\
3.75	0.420977663341156\\
};
\addplot [color=blue, line width=0.25mm, forget plot]
  table[row sep=crcr]{%
4.75	0.408502457853301\\
4.75	0.769962145581353\\
5.25	0.769962145581353\\
5.25	0.408502457853301\\
4.75	0.408502457853301\\
};
\addplot [color=red, line width=0.25mm, forget plot]
  table[row sep=crcr]{%
0.75	0.817275389949671\\
1.25	0.817275389949671\\
};
\addplot [color=red, line width=0.25mm, forget plot]
  table[row sep=crcr]{%
1.75	0.753379830086935\\
2.25	0.753379830086935\\
};
\addplot [color=red, line width=0.25mm, forget plot]
  table[row sep=crcr]{%
2.75	0.728268445122613\\
3.25	0.728268445122613\\
};
\addplot [color=red, line width=0.25mm, forget plot]
  table[row sep=crcr]{%
3.75	0.615171756557372\\
4.25	0.615171756557372\\
};
\addplot [color=red, line width=0.25mm, forget plot]
  table[row sep=crcr]{%
4.75	0.662475970608626\\
5.25	0.662475970608626\\
};
\end{axis}
\end{tikzpicture}%
}} &
  \subfloat[]{
    \resizebox{0.33\textwidth}{!}{
%
%
\begin{tikzpicture}
\definecolor{blue}{RGB}{76,100,135}
\definecolor{red}{RGB}{153,0,0}
\begin{axis}[%
width=4.634in,
height=3.612in,
at={(0.777in,0.487in)},
xmin=0.5,
xmax=5.5,
xtick={1,2,3,4, 5},
xticklabels={s-IK,s-IK2,m-IK,IK, e-IK},
ymin=0.0967363899401256,
ymax=2.25047949341658,
ytick={ 1, 2},
yticklabels={1, 2},
ylabel style={font=\color{white!15!black}},
ylabel={$\sigma{}_{\text{min}}$},
ylabel near ticks,
ymajorgrids,
scale=0.5,
axis background/.style={fill=white},
title style={font=\bfseries},
title={6 DOF},
legend style={legend cell align=left, align=left, draw=white!15!black}
]
\addplot [color=black, dashed, line width=0.25mm, forget plot]
  table[row sep=crcr]{%
1	1.84148479411149\\
1	1.96044552432128\\
};
\addplot [color=black, dashed, line width=0.25mm, forget plot]
  table[row sep=crcr]{%
2	1.63645470123703\\
2	1.72513534962181\\
};
\addplot [color=black, dashed, line width=0.25mm, forget plot]
  table[row sep=crcr]{%
3	1.65269257993084\\
3	1.72583284377362\\
};
\addplot [color=black, dashed, line width=0.25mm, forget plot]
  table[row sep=crcr]{%
4	1.13469634958286\\
4	1.63204232711052\\
};
\addplot [color=black, dashed, line width=0.25mm, forget plot]
  table[row sep=crcr]{%
5	1.18139505838962\\
5	1.24606736717049\\
};
\addplot [color=black, dashed, line width=0.25mm, forget plot]
  table[row sep=crcr]{%
1	1.08556350787759\\
1	1.41283326205588\\
};
\addplot [color=black, dashed, line width=0.25mm, forget plot]
  table[row sep=crcr]{%
2	1.00813426434206\\
2	1.27810681267219\\
};
\addplot [color=black, dashed, line width=0.25mm, forget plot]
  table[row sep=crcr]{%
3	0.997809178299127\\
3	1.28663920207722\\
};
\addplot [color=black, dashed, line width=0.25mm, forget plot]
  table[row sep=crcr]{%
4	0.236687122621615\\
4	0.744462299602703\\
};
\addplot [color=black, dashed, line width=0.25mm, forget plot]
  table[row sep=crcr]{%
5	0.679066280548257\\
5	0.887658490891164\\
};
\addplot [color=black, line width=0.25mm, forget plot]
  table[row sep=crcr]{%
0.875	1.96044552432128\\
1.125	1.96044552432128\\
};
\addplot [color=black, line width=0.25mm, forget plot]
  table[row sep=crcr]{%
1.875	1.72513534962181\\
2.125	1.72513534962181\\
};
\addplot [color=black, line width=0.25mm, forget plot]
  table[row sep=crcr]{%
2.875	1.72583284377362\\
3.125	1.72583284377362\\
};
\addplot [color=black, line width=0.25mm, forget plot]
  table[row sep=crcr]{%
3.875	1.63204232711052\\
4.125	1.63204232711052\\
};
\addplot [color=black, line width=0.25mm, forget plot]
  table[row sep=crcr]{%
4.875	1.24606736717049\\
5.125	1.24606736717049\\
};
\addplot [color=black, line width=0.25mm, forget plot]
  table[row sep=crcr]{%
0.875	1.08556350787759\\
1.125	1.08556350787759\\
};
\addplot [color=black, line width=0.25mm, forget plot]
  table[row sep=crcr]{%
1.875	1.00813426434206\\
2.125	1.00813426434206\\
};
\addplot [color=black, line width=0.25mm, forget plot]
  table[row sep=crcr]{%
2.875	0.997809178299127\\
3.125	0.997809178299127\\
};
\addplot [color=black, line width=0.25mm, forget plot]
  table[row sep=crcr]{%
3.875	0.236687122621615\\
4.125	0.236687122621615\\
};
\addplot [color=black, line width=0.25mm, forget plot]
  table[row sep=crcr]{%
4.875	0.679066280548257\\
5.125	0.679066280548257\\
};
\addplot [color=blue, line width=0.25mm, forget plot]
  table[row sep=crcr]{%
0.75	1.41283326205588\\
0.75	1.84148479411149\\
1.25	1.84148479411149\\
1.25	1.41283326205588\\
0.75	1.41283326205588\\
};
\addplot [color=blue, line width=0.25mm, forget plot]
  table[row sep=crcr]{%
1.75	1.27810681267219\\
1.75	1.63645470123703\\
2.25	1.63645470123703\\
2.25	1.27810681267219\\
1.75	1.27810681267219\\
};
\addplot [color=blue, line width=0.25mm, forget plot]
  table[row sep=crcr]{%
2.75	1.28663920207722\\
2.75	1.65269257993084\\
3.25	1.65269257993084\\
3.25	1.28663920207722\\
2.75	1.28663920207722\\
};
\addplot [color=blue, line width=0.25mm, forget plot]
  table[row sep=crcr]{%
3.75	0.744462299602703\\
3.75	1.13469634958286\\
4.25	1.13469634958286\\
4.25	0.744462299602703\\
3.75	0.744462299602703\\
};
\addplot [color=blue, line width=0.25mm, forget plot]
  table[row sep=crcr]{%
4.75	0.887658490891164\\
4.75	1.18139505838962\\
5.25	1.18139505838962\\
5.25	0.887658490891164\\
4.75	0.887658490891164\\
};
\addplot [color=red, line width=0.25mm, forget plot]
  table[row sep=crcr]{%
0.75	1.6722789495416\\
1.25	1.6722789495416\\
};
\addplot [color=red, line width=0.25mm, forget plot]
  table[row sep=crcr]{%
1.75	1.4959343890566\\
2.25	1.4959343890566\\
};
\addplot [color=red, line width=0.25mm, forget plot]
  table[row sep=crcr]{%
2.75	1.52091018759452\\
3.25	1.52091018759452\\
};
\addplot [color=red, line width=0.25mm, forget plot]
  table[row sep=crcr]{%
3.75	0.958007281265328\\
4.25	0.958007281265328\\
};
\addplot [color=red, line width=0.25mm, forget plot]
  table[row sep=crcr]{%
4.75	0.939548080082356\\
5.25	0.939548080082356\\
};
\end{axis}

\end{tikzpicture}%
}} &
  \subfloat[]{
    \resizebox{0.33\textwidth}{!}{
%
%
\begin{tikzpicture}
\definecolor{blue}{RGB}{76,100,135}
\definecolor{red}{RGB}{153,0,0}
\begin{axis}[%
width=4.568in,
height=3.603in,
at={(0.766in,0.486in)},
xmin=0.5,
xmax=5.5,
xtick={1,2,3,4,5},
xticklabels={s-IK,s-IK2,m-IK,IK, e-IK},
ymin=0,
ymax=4.49942728613223,
ytick={ 2, 4},
yticklabels={2, 4},
ylabel style={font=\color{white!15!black}},
ylabel={$\sigma{}_{\text{min}}$},
ylabel near ticks,
ymajorgrids,
scale=0.5,
axis background/.style={fill=white},
title style={font=\bfseries},
title={9 DOF},
legend style={legend cell align=left, align=left, draw=white!15!black}
]
\addplot [color=black, dashed, line width=0.25mm, forget plot]
  table[row sep=crcr]{%
1	3.76654160027719\\
1	3.94136764521233\\
};
\addplot [color=black, dashed, line width=0.25mm, forget plot]
  table[row sep=crcr]{%
2	3.24506879738002\\
2	3.45801774808808\\
};
\addplot [color=black, dashed, line width=0.25mm, forget plot]
  table[row sep=crcr]{%
3	3.31699097481201\\
3	3.46290428082503\\
};
\addplot [color=black, dashed, line width=0.25mm, forget plot]
  table[row sep=crcr]{%
4	1.79640395952656\\
4	2.56879458045882\\
};
\addplot [color=black, dashed, line width=0.25mm, forget plot]
  table[row sep=crcr]{%
5	2.94123962267591\\
5	3.11984290903765\\
};
\addplot [color=black, dashed, line width=0.25mm, forget plot]
  table[row sep=crcr]{%
1	2.46994138556031\\
1	3.20331027613889\\
};
\addplot [color=black, dashed, line width=0.25mm, forget plot]
  table[row sep=crcr]{%
2	2.08670620912744\\
2	2.76290715701044\\
};
\addplot [color=black, dashed, line width=0.25mm, forget plot]
  table[row sep=crcr]{%
3	2.31795165814465\\
3	2.87399928081068\\
};
\addplot [color=black, dashed, line width=0.25mm, forget plot]
  table[row sep=crcr]{%
4	0.713606310313453\\
4	1.25140846261399\\
};
\addplot [color=black, dashed, line width=0.25mm, forget plot]
  table[row sep=crcr]{%
5	1.75339251930155\\
5	2.46231904980744\\
};
\addplot [color=black, line width=0.25mm, forget plot]
  table[row sep=crcr]{%
0.875	3.94136764521233\\
1.125	3.94136764521233\\
};
\addplot [color=black, line width=0.25mm, forget plot]
  table[row sep=crcr]{%
1.875	3.45801774808808\\
2.125	3.45801774808808\\
};
\addplot [color=black, line width=0.25mm, forget plot]
  table[row sep=crcr]{%
2.875	3.46290428082503\\
3.125	3.46290428082503\\
};
\addplot [color=black, line width=0.25mm, forget plot]
  table[row sep=crcr]{%
3.875	2.56879458045882\\
4.125	2.56879458045882\\
};
\addplot [color=black, line width=0.25mm, forget plot]
  table[row sep=crcr]{%
4.875	3.11984290903765\\
5.125	3.11984290903765\\
};
\addplot [color=black, line width=0.25mm, forget plot]
  table[row sep=crcr]{%
0.875	2.46994138556031\\
1.125	2.46994138556031\\
};
\addplot [color=black, line width=0.25mm, forget plot]
  table[row sep=crcr]{%
1.875	2.08670620912744\\
2.125	2.08670620912744\\
};
\addplot [color=black, line width=0.25mm, forget plot]
  table[row sep=crcr]{%
2.875	2.31795165814465\\
3.125	2.31795165814465\\
};
\addplot [color=black, line width=0.25mm, forget plot]
  table[row sep=crcr]{%
3.875	0.713606310313453\\
4.125	0.713606310313453\\
};
\addplot [color=black, line width=0.25mm, forget plot]
  table[row sep=crcr]{%
4.875	1.75339251930155\\
5.125	1.75339251930155\\
};
\addplot [color=blue, line width=0.25mm, forget plot]
  table[row sep=crcr]{%
0.75	3.20331027613889\\
0.75	3.76654160027719\\
1.25	3.76654160027719\\
1.25	3.20331027613889\\
0.75	3.20331027613889\\
};
\addplot [color=blue, line width=0.25mm, forget plot]
  table[row sep=crcr]{%
1.75	2.76290715701044\\
1.75	3.24506879738002\\
2.25	3.24506879738002\\
2.25	2.76290715701044\\
1.75	2.76290715701044\\
};
\addplot [color=blue, line width=0.25mm, forget plot]
  table[row sep=crcr]{%
2.75	2.87399928081068\\
2.75	3.31699097481201\\
3.25	3.31699097481201\\
3.25	2.87399928081068\\
2.75	2.87399928081068\\
};
\addplot [color=blue, line width=0.25mm, forget plot]
  table[row sep=crcr]{%
3.75	1.25140846261399\\
3.75	1.79640395952656\\
4.25	1.79640395952656\\
4.25	1.25140846261399\\
3.75	1.25140846261399\\
};
\addplot [color=blue, line width=0.25mm, forget plot]
  table[row sep=crcr]{%
4.75	2.46231904980744\\
4.75	2.94123962267591\\
5.25	2.94123962267591\\
5.25	2.46231904980744\\
4.75	2.46231904980744\\
};
\addplot [color=red, line width=0.25mm, forget plot]
  table[row sep=crcr]{%
0.75	3.5385388285547\\
1.25	3.5385388285547\\
};
\addplot [color=red, line width=0.25mm, forget plot]
  table[row sep=crcr]{%
1.75	3.05161899253248\\
2.25	3.05161899253248\\
};
\addplot [color=red, line width=0.25mm, forget plot]
  table[row sep=crcr]{%
2.75	3.11337722620248\\
3.25	3.11337722620248\\
};
\addplot [color=red, line width=0.25mm, forget plot]
  table[row sep=crcr]{%
3.75	1.50637049185261\\
4.25	1.50637049185261\\
};
\addplot [color=red, line width=0.25mm, forget plot]
  table[row sep=crcr]{%
4.75	2.76728798273472\\
5.25	2.76728798273472\\
};
\end{axis}

\end{tikzpicture}%
}} \\[-4ex]
  \subfloat[]{
    \resizebox{0.33\textwidth}{!}{
%
%
\begin{tikzpicture}
\definecolor{blue}{RGB}{76,100,135}
\definecolor{red}{RGB}{153,0,0}
\begin{axis}[%
width=4.634in,
height=3.612in,
at={(0.777in,0.487in)},
unbounded coords=jump,
xmin=0.5,
xmax=5.5,
xtick={1,2,3,4,5},
xticklabels={s-IK,s-IK2,m-IK,IK, e-IK},
ymin=1.02077870131498,
ymax=3.36442052152183,
ytick={ 2, 3},
yticklabels={2, 3},
ylabel style={font=\color{white!15!black}},
ylabel={$\sigma{}_{\text{max}}$},
ymajorgrids,
ylabel near ticks,
scale=0.5,
axis background/.style={fill=white},
legend style={legend cell align=left, align=left, draw=white!15!black}
]
\addplot [color=black, dashed, line width=0.25mm, forget plot]
  table[row sep=crcr]{%
1	2.20089521009681\\
1	3.0730847221524\\
};
\addplot [color=black, dashed, line width=0.25mm, forget plot]
  table[row sep=crcr]{%
2	2.27997990270494\\
2	3.07456800046276\\
};
\addplot [color=black, dashed, line width=0.25mm, forget plot]
  table[row sep=crcr]{%
3	2.31265384727008\\
3	3.07456279753737\\
};
\addplot [color=black, dashed, line width=0.25mm, forget plot]
  table[row sep=crcr]{%
4	2.23097799342863\\
4	3.07376410606504\\
};
\addplot [color=black, dashed, line width=0.25mm, forget plot]
  table[row sep=crcr]{%
5	2.39449192368398\\
5	3.07529165006593\\
};
\addplot [color=black, dashed, line width=0.25mm, forget plot]
  table[row sep=crcr]{%
1	1.11941726981624\\
1	1.3665649418186\\
};
\addplot [color=black, dashed, line width=0.25mm, forget plot]
  table[row sep=crcr]{%
2	1.34714894838474\\
2	1.51197353351978\\
};
\addplot [color=black, dashed, line width=0.25mm, forget plot]
  table[row sep=crcr]{%
3	1.15844017907341\\
3	1.52585977550049\\
};
\addplot [color=black, dashed, line width=0.25mm, forget plot]
  table[row sep=crcr]{%
4	1.13914634437206\\
4	1.40249696841098\\
};
\addplot [color=black, dashed, line width=0.25mm, forget plot]
  table[row sep=crcr]{%
5	1.15198791879192\\
5	1.88976969355926\\
};
\addplot [color=black, line width=0.25mm, forget plot]
  table[row sep=crcr]{%
0.875	3.0730847221524\\
1.125	3.0730847221524\\
};
\addplot [color=black, line width=0.25mm, forget plot]
  table[row sep=crcr]{%
1.875	3.07456800046276\\
2.125	3.07456800046276\\
};
\addplot [color=black, line width=0.25mm, forget plot]
  table[row sep=crcr]{%
2.875	3.07456279753737\\
3.125	3.07456279753737\\
};
\addplot [color=black, line width=0.25mm, forget plot]
  table[row sep=crcr]{%
3.875	3.07376410606504\\
4.125	3.07376410606504\\
};
\addplot [color=black, line width=0.25mm, forget plot]
  table[row sep=crcr]{%
4.875	3.07529165006593\\
5.125	3.07529165006593\\
};
\addplot [color=black, line width=0.25mm, forget plot]
  table[row sep=crcr]{%
0.875	1.11941726981624\\
1.125	1.11941726981624\\
};
\addplot [color=black, line width=0.25mm, forget plot]
  table[row sep=crcr]{%
1.875	1.34714894838474\\
2.125	1.34714894838474\\
};
\addplot [color=black, line width=0.25mm, forget plot]
  table[row sep=crcr]{%
2.875	1.15844017907341\\
3.125	1.15844017907341\\
};
\addplot [color=black, line width=0.25mm, forget plot]
  table[row sep=crcr]{%
3.875	1.13914634437206\\
4.125	1.13914634437206\\
};
\addplot [color=black, line width=0.25mm, forget plot]
  table[row sep=crcr]{%
4.875	1.15198791879192\\
5.125	1.15198791879192\\
};
\addplot [color=blue, line width=0.25mm, forget plot]
  table[row sep=crcr]{%
0.75	1.3665649418186\\
0.75	2.20089521009681\\
1.25	2.20089521009681\\
1.25	1.3665649418186\\
0.75	1.3665649418186\\
};
\addplot [color=blue, line width=0.25mm, forget plot]
  table[row sep=crcr]{%
1.75	1.51197353351978\\
1.75	2.27997990270494\\
2.25	2.27997990270494\\
2.25	1.51197353351978\\
1.75	1.51197353351978\\
};
\addplot [color=blue, line width=0.25mm, forget plot]
  table[row sep=crcr]{%
2.75	1.52585977550049\\
2.75	2.31265384727008\\
3.25	2.31265384727008\\
3.25	1.52585977550049\\
2.75	1.52585977550049\\
};
\addplot [color=blue, line width=0.25mm, forget plot]
  table[row sep=crcr]{%
3.75	1.40249696841098\\
3.75	2.23097799342863\\
4.25	2.23097799342863\\
4.25	1.40249696841098\\
3.75	1.40249696841098\\
};
\addplot [color=blue, line width=0.25mm, forget plot]
  table[row sep=crcr]{%
4.75	1.88976969355926\\
4.75	2.39449192368398\\
5.25	2.39449192368398\\
5.25	1.88976969355926\\
4.75	1.88976969355926\\
};
\addplot [color=red, line width=0.25mm, forget plot]
  table[row sep=crcr]{%
0.75	1.64582249129141\\
1.25	1.64582249129141\\
};
\addplot [color=red, line width=0.25mm, forget plot]
  table[row sep=crcr]{%
1.75	1.88547556297999\\
2.25	1.88547556297999\\
};
\addplot [color=red, line width=0.25mm, forget plot]
  table[row sep=crcr]{%
2.75	1.84333226498963\\
3.25	1.84333226498963\\
};
\addplot [color=red, line width=0.25mm, forget plot]
  table[row sep=crcr]{%
3.75	1.84524287705235\\
4.25	1.84524287705235\\
};
\addplot [color=red, line width=0.25mm, forget plot]
  table[row sep=crcr]{%
4.75	2.06450874245023\\
5.25	2.06450874245023\\
};
\end{axis}
\end{tikzpicture}%
}} &
  \subfloat[]{
    \resizebox{0.33\textwidth}{!}{
%
%
\begin{tikzpicture}
\definecolor{blue}{RGB}{76,100,135}
\definecolor{red}{RGB}{153,0,0}
\begin{axis}[%
width=4.634in,
height=3.612in,
at={(0.777in,0.487in)},
unbounded coords=jump,
xmin=0.5,
xmax=5.5,
xtick={1,2,3,4,5},
xticklabels={s-IK,s-IK2,m-IK,IK,e-IK},
ymin=1.02584498842977,
ymax=5.73359142365963,
ytick={ 3, 5},
yticklabels={3, 5},
ylabel style={font=\color{white!15!black}},
ylabel={$\sigma{}_{\text{max}}$},
ylabel near ticks,
ymajorgrids,
scale=0.5,
axis background/.style={fill=white},
legend style={legend cell align=left, align=left, draw=white!15!black}
]
\addplot [color=black, dashed, line width=0.25mm, forget plot]
  table[row sep=crcr]{%
1	3.35336476120956\\
1	4.86984245818414\\
};
\addplot [color=black, dashed, line width=0.25mm, forget plot]
  table[row sep=crcr]{%
2	3.70899013716819\\
2	4.98721268925112\\
};
\addplot [color=black, dashed, line width=0.25mm, forget plot]
  table[row sep=crcr]{%
3	3.77326808526417\\
3	5.02929835683418\\
};
\addplot [color=black, dashed, line width=0.25mm, forget plot]
  table[row sep=crcr]{%
4	3.93458331131311\\
4	5.07649395182143\\
};
\addplot [color=black, dashed, line width=0.25mm, forget plot]
  table[row sep=crcr]{%
5	4.45773018341364\\
5	5.24122174143907\\
};
\addplot [color=black, dashed, line width=0.25mm, forget plot]
  table[row sep=crcr]{%
1	1.69034992089572\\
1	2.34066626166597\\
};
\addplot [color=black, dashed, line width=0.25mm, forget plot]
  table[row sep=crcr]{%
2	2.16540969593411\\
2	2.8465558012276\\
};
\addplot [color=black, dashed, line width=0.25mm, forget plot]
  table[row sep=crcr]{%
3	2.14447984372695\\
3	2.85596026459146\\
};
\addplot [color=black, dashed, line width=0.25mm, forget plot]
  table[row sep=crcr]{%
4	1.27545680434936\\
4	2.09834772274873\\
};
\addplot [color=black, dashed, line width=0.25mm, forget plot]
  table[row sep=crcr]{%
5	2.55516989410199\\
5	3.59697805406507\\
};
\addplot [color=black, line width=0.25mm, forget plot]
  table[row sep=crcr]{%
0.875	4.86984245818414\\
1.125	4.86984245818414\\
};
\addplot [color=black, line width=0.25mm, forget plot]
  table[row sep=crcr]{%
1.875	4.98721268925112\\
2.125	4.98721268925112\\
};
\addplot [color=black, line width=0.25mm, forget plot]
  table[row sep=crcr]{%
2.875	5.02929835683418\\
3.125	5.02929835683418\\
};
\addplot [color=black, line width=0.25mm, forget plot]
  table[row sep=crcr]{%
3.875	5.07649395182143\\
4.125	5.07649395182143\\
};
\addplot [color=black, line width=0.25mm, forget plot]
  table[row sep=crcr]{%
4.875	5.24122174143907\\
5.125	5.24122174143907\\
};
\addplot [color=black, line width=0.25mm, forget plot]
  table[row sep=crcr]{%
0.875	1.69034992089572\\
1.125	1.69034992089572\\
};
\addplot [color=black, line width=0.25mm, forget plot]
  table[row sep=crcr]{%
1.875	2.16540969593411\\
2.125	2.16540969593411\\
};
\addplot [color=black, line width=0.25mm, forget plot]
  table[row sep=crcr]{%
2.875	2.14447984372695\\
3.125	2.14447984372695\\
};
\addplot [color=black, line width=0.25mm, forget plot]
  table[row sep=crcr]{%
3.875	1.27545680434936\\
4.125	1.27545680434936\\
};
\addplot [color=black, line width=0.25mm, forget plot]
  table[row sep=crcr]{%
4.875	2.55516989410199\\
5.125	2.55516989410199\\
};
\addplot [color=blue, line width=0.25mm, forget plot]
  table[row sep=crcr]{%
0.75	2.34066626166597\\
0.75	3.35336476120956\\
1.25	3.35336476120956\\
1.25	2.34066626166597\\
0.75	2.34066626166597\\
};
\addplot [color=blue, line width=0.25mm, forget plot]
  table[row sep=crcr]{%
1.75	2.8465558012276\\
1.75	3.70899013716819\\
2.25	3.70899013716819\\
2.25	2.8465558012276\\
1.75	2.8465558012276\\
};
\addplot [color=blue, line width=0.25mm, forget plot]
  table[row sep=crcr]{%
2.75	2.85596026459146\\
2.75	3.77326808526417\\
3.25	3.77326808526417\\
3.25	2.85596026459146\\
2.75	2.85596026459146\\
};
\addplot [color=blue, line width=0.25mm, forget plot]
  table[row sep=crcr]{%
3.75	2.09834772274873\\
3.75	3.93458331131311\\
4.25	3.93458331131311\\
4.25	2.09834772274873\\
3.75	2.09834772274873\\
};
\addplot [color=blue, line width=0.25mm, forget plot]
  table[row sep=crcr]{%
4.75	3.59697805406507\\
4.75	4.45773018341364\\
5.25	4.45773018341364\\
5.25	3.59697805406507\\
4.75	3.59697805406507\\
};
\addplot [color=red, line width=0.25mm, forget plot]
  table[row sep=crcr]{%
0.75	2.54115344642591\\
1.25	2.54115344642591\\
};
\addplot [color=red, line width=0.25mm, forget plot]
  table[row sep=crcr]{%
1.75	3.1666367126392\\
2.25	3.1666367126392\\
};
\addplot [color=red, line width=0.25mm, forget plot]
  table[row sep=crcr]{%
2.75	3.17186752457941\\
3.25	3.17186752457941\\
};
\addplot [color=red, line width=0.25mm, forget plot]
  table[row sep=crcr]{%
3.75	2.9165638457184\\
4.25	2.9165638457184\\
};
\addplot [color=red, line width=0.25mm, forget plot]
  table[row sep=crcr]{%
4.75	4.01930396111373\\
5.25	4.01930396111373\\
};
\end{axis}

\end{tikzpicture}%
}} &
  \subfloat[]{
    \resizebox{0.33\textwidth}{!}{
%
%
\begin{tikzpicture}
\definecolor{blue}{RGB}{76,100,135}
\definecolor{red}{RGB}{153,0,0}
\begin{axis}[%
width=4.568in,
height=3.603in,
at={(0.766in,0.486in)},
xmin=0.5,
xmax=5.5,
xtick={1,2,3,4,5},
xticklabels={s-IK,s-IK2,m-IK,IK,e-IK},
ymin=1.57579639935418,
ymax=11.5053424683116,
ytick={ 6, 10},
yticklabels={6, 10},
ylabel style={font=\color{white!15!black}},
ylabel={$\sigma{}_{\text{max}}$},
ylabel near ticks,
ymajorgrids,
scale=0.5,
axis background/.style={fill=white},
legend style={legend cell align=left, align=left, draw=white!15!black}
]
\addplot [color=black, dashed, line width=0.25mm, forget plot]
  table[row sep=crcr]{%
1	6.05426902150847\\
1	7.69087404148088\\
};
\addplot [color=black, dashed, line width=0.25mm, forget plot]
  table[row sep=crcr]{%
2	8.04363799819786\\
2	10.2791383945382\\
};
\addplot [color=black, dashed, line width=0.25mm, forget plot]
  table[row sep=crcr]{%
3	8.02578942424537\\
3	10.1118363774653\\
};
\addplot [color=black, dashed, line width=0.25mm, forget plot]
  table[row sep=crcr]{%
4	6.86764429439074\\
4	11.2592672252808\\
};
\addplot [color=black, dashed, line width=0.25mm, forget plot]
  table[row sep=crcr]{%
5	8.802734636247\\
5	10.9815589925106\\
};
\addplot [color=black, dashed, line width=0.25mm, forget plot]
  table[row sep=crcr]{%
1	3.78230478893113\\
1	4.94321571954651\\
};
\addplot [color=black, dashed, line width=0.25mm, forget plot]
  table[row sep=crcr]{%
2	4.82426420241018\\
2	6.48927244249518\\
};
\addplot [color=black, dashed, line width=0.25mm, forget plot]
  table[row sep=crcr]{%
3	4.81247775687421\\
3	6.54743461474675\\
};
\addplot [color=black, dashed, line width=0.25mm, forget plot]
  table[row sep=crcr]{%
4	1.95022070041891\\
4	3.68050627841935\\
};
\addplot [color=black, dashed, line width=0.25mm, forget plot]
  table[row sep=crcr]{%
5	6.07595815794114\\
5	7.3045026950807\\
};
\addplot [color=black, line width=0.25mm, forget plot]
  table[row sep=crcr]{%
0.875	7.69087404148088\\
1.125	7.69087404148088\\
};
\addplot [color=black, line width=0.25mm, forget plot]
  table[row sep=crcr]{%
1.875	10.2791383945382\\
2.125	10.2791383945382\\
};
\addplot [color=black, line width=0.25mm, forget plot]
  table[row sep=crcr]{%
2.875	10.1118363774653\\
3.125	10.1118363774653\\
};
\addplot [color=black, line width=0.25mm, forget plot]
  table[row sep=crcr]{%
3.875	11.2592672252808\\
4.125	11.2592672252808\\
};
\addplot [color=black, line width=0.25mm, forget plot]
  table[row sep=crcr]{%
4.875	10.9815589925106\\
5.125	10.9815589925106\\
};
\addplot [color=black, line width=0.25mm, forget plot]
  table[row sep=crcr]{%
0.875	3.78230478893113\\
1.125	3.78230478893113\\
};
\addplot [color=black, line width=0.25mm, forget plot]
  table[row sep=crcr]{%
1.875	4.82426420241018\\
2.125	4.82426420241018\\
};
\addplot [color=black, line width=0.25mm, forget plot]
  table[row sep=crcr]{%
2.875	4.81247775687421\\
3.125	4.81247775687421\\
};
\addplot [color=black, line width=0.25mm, forget plot]
  table[row sep=crcr]{%
3.875	1.95022070041891\\
4.125	1.95022070041891\\
};
\addplot [color=black, line width=0.25mm, forget plot]
  table[row sep=crcr]{%
4.875	6.07595815794114\\
5.125	6.07595815794114\\
};
\addplot [color=blue, line width=0.25mm, forget plot]
  table[row sep=crcr]{%
0.75	4.94321571954651\\
0.75	6.05426902150847\\
1.25	6.05426902150847\\
1.25	4.94321571954651\\
0.75	4.94321571954651\\
};
\addplot [color=blue, line width=0.25mm, forget plot]
  table[row sep=crcr]{%
1.75	6.48927244249518\\
1.75	8.04363799819786\\
2.25	8.04363799819786\\
2.25	6.48927244249518\\
1.75	6.48927244249518\\
};
\addplot [color=blue, line width=0.25mm, forget plot]
  table[row sep=crcr]{%
2.75	6.54743461474675\\
2.75	8.02578942424537\\
3.25	8.02578942424537\\
3.25	6.54743461474675\\
2.75	6.54743461474675\\
};
\addplot [color=blue, line width=0.25mm, forget plot]
  table[row sep=crcr]{%
3.75	3.68050627841935\\
3.75	6.86764429439074\\
4.25	6.86764429439074\\
4.25	3.68050627841935\\
3.75	3.68050627841935\\
};
\addplot [color=blue, line width=0.25mm, forget plot]
  table[row sep=crcr]{%
4.75	7.3045026950807\\
4.75	8.802734636247\\
5.25	8.802734636247\\
5.25	7.3045026950807\\
4.75	7.3045026950807\\
};
\addplot [color=red, line width=0.25mm, forget plot]
  table[row sep=crcr]{%
0.75	5.46863429025012\\
1.25	5.46863429025012\\
};
\addplot [color=red, line width=0.25mm, forget plot]
  table[row sep=crcr]{%
1.75	7.29896431268372\\
2.25	7.29896431268372\\
};
\addplot [color=red, line width=0.25mm, forget plot]
  table[row sep=crcr]{%
2.75	7.32521930874011\\
3.25	7.32521930874011\\
};
\addplot [color=red, line width=0.25mm, forget plot]
  table[row sep=crcr]{%
3.75	5.38444066337042\\
4.25	5.38444066337042\\
};
\addplot [color=red, line width=0.25mm, forget plot]
  table[row sep=crcr]{%
4.75	8.03631450000728\\
5.25	8.03631450000728\\
};
\end{axis}

\end{tikzpicture}%
}} 
  \end{tabular}
  }
  \caption{Results of solving 200 random inverse kinematics (IK) problems; each column corresponds to IK solutions for a planar manipulator with a different number of DoF. The plots in the top row show the minimal singular values $\sigma_{\text{min}}$ of the manipulator Jacobian in the final configuration, while the plots in the bottom row show the maximal singular values $\sigma_{\text{max}}$. The two leftmost boxes in each plot, labeled s-IK and s-IK2, represent our method with for different choices of $\bSigma$. The box labeled m-IK corresponds to the method in \cite{dufour2017integrating}, while the box labeled IK shows the results without optimizing for singularity avoidance. Finally, the box labeled e-IK shows the results obtained when using a Euclidean metric.}\label{fig:exp1}
\end{figure*}

\begin{figure*}[!t]
  \captionsetup[subfloat]{position=top,labelformat=empty}
  \centering
  \hspace*{-0.6cm}
    \resizebox{\textwidth}{!}{
  \begin{tabular}{@{}ccc@{}}
  \subfloat[]{
    \resizebox{0.33\textwidth}{!}{
%
%
\begin{tikzpicture}
\definecolor{blue}{RGB}{76,100,135}
\definecolor{red}{RGB}{153,0,0}

\begin{axis}[%
width=4.634in,
height=3.612in,
at={(0.777in,0.487in)},
unbounded coords=jump,
xmin=0.5,
xmax=5.5,
xtick={1,2,3,4,5},
xticklabels={s-IK,s-IK2,m-IK,IK, e-IK},
ymin=0.00000,
ymax=0.60,
ytick={ 0.2, 0.4},
yticklabels={0.2, 0.4},
ylabel style={font=\color{white!15!black}},
ylabel={$\sigma{}_{\text{min}}$},
ylabel near ticks,
ymajorgrids,
scale=0.5,
axis background/.style={fill=white},
title style={font=\bfseries},
title={UR-10},
legend style={legend cell align=left, align=left, draw=white!15!black}
]
\addplot [color=black, dashed, line width=0.25mm, forget plot]
  table[row sep=crcr]{%
1	0.383955752791766\\
1	0.452039715172139\\
};
\addplot [color=black, dashed, line width=0.25mm, forget plot]
  table[row sep=crcr]{%
2	0.378686312099035\\
2	0.443639517130951\\
};
\addplot [color=black, dashed, line width=0.25mm, forget plot]
  table[row sep=crcr]{%
3	0.376685696748964\\
3	0.444105652808365\\
};
\addplot [color=black, dashed, line width=0.25mm, forget plot]
  table[row sep=crcr]{%
4	0.32684052298746\\
4	0.446270275602577\\
};
\addplot [color=black, dashed, line width=0.25mm, forget plot]
  table[row sep=crcr]{%
5	0.367725148307756\\
5	0.442207390402739\\
};
\addplot [color=black, dashed, line width=0.25mm, forget plot]
  table[row sep=crcr]{%
1	0.000270575914893843\\
1	0.182471936646954\\
};
\addplot [color=black, dashed, line width=0.25mm, forget plot]
  table[row sep=crcr]{%
2	0.000513251948309454\\
2	0.185837685960821\\
};
\addplot [color=black, dashed, line width=0.25mm, forget plot]
  table[row sep=crcr]{%
3	0.000687902534064449\\
3	0.176806511986295\\
};
\addplot [color=black, dashed, line width=0.25mm, forget plot]
  table[row sep=crcr]{%
4	0.000407015156399139\\
4	0.124915806152405\\
};
\addplot [color=black, dashed, line width=0.25mm, forget plot]
  table[row sep=crcr]{%
5	0.000476298678930936\\
5	0.158082177430966\\
};
\addplot [color=black, line width=0.25mm, forget plot]
  table[row sep=crcr]{%
0.875	0.452039715172139\\
1.125	0.452039715172139\\
};
\addplot [color=black, line width=0.25mm, forget plot]
  table[row sep=crcr]{%
1.875	0.443639517130951\\
2.125	0.443639517130951\\
};
\addplot [color=black, line width=0.25mm, forget plot]
  table[row sep=crcr]{%
2.875	0.444105652808365\\
3.125	0.444105652808365\\
};
\addplot [color=black, line width=0.25mm, forget plot]
  table[row sep=crcr]{%
3.875	0.446270275602577\\
4.125	0.446270275602577\\
};
\addplot [color=black, line width=0.25mm, forget plot]
  table[row sep=crcr]{%
4.875	0.442207390402739\\
5.125	0.442207390402739\\
};
\addplot [color=black, line width=0.25mm, forget plot]
  table[row sep=crcr]{%
0.875	0.000270575914893843\\
1.125	0.000270575914893843\\
};
\addplot [color=black, line width=0.25mm, forget plot]
  table[row sep=crcr]{%
1.875	0.000513251948309454\\
2.125	0.000513251948309454\\
};
\addplot [color=black, line width=0.25mm, forget plot]
  table[row sep=crcr]{%
2.875	0.000687902534064449\\
3.125	0.000687902534064449\\
};
\addplot [color=black, line width=0.25mm, forget plot]
  table[row sep=crcr]{%
3.875	0.000407015156399139\\
4.125	0.000407015156399139\\
};
\addplot [color=black, line width=0.25mm, forget plot]
  table[row sep=crcr]{%
4.875	0.000476298678930936\\
5.125	0.000476298678930936\\
};
\addplot [color=blue, line width=0.25mm, forget plot]
  table[row sep=crcr]{%
0.75	0.182471936646954\\
0.75	0.383955752791766\\
1.25	0.383955752791766\\
1.25	0.182471936646954\\
0.75	0.182471936646954\\
};
\addplot [color=blue, line width=0.25mm, forget plot]
  table[row sep=crcr]{%
1.75	0.185837685960821\\
1.75	0.378686312099035\\
2.25	0.378686312099035\\
2.25	0.185837685960821\\
1.75	0.185837685960821\\
};
\addplot [color=blue, line width=0.25mm, forget plot]
  table[row sep=crcr]{%
2.75	0.176806511986295\\
2.75	0.376685696748964\\
3.25	0.376685696748964\\
3.25	0.176806511986295\\
2.75	0.176806511986295\\
};
\addplot [color=blue, line width=0.25mm, forget plot]
  table[row sep=crcr]{%
3.75	0.124915806152405\\
3.75	0.32684052298746\\
4.25	0.32684052298746\\
4.25	0.124915806152405\\
3.75	0.124915806152405\\
};
\addplot [color=blue, line width=0.25mm, forget plot]
  table[row sep=crcr]{%
4.75	0.158082177430966\\
4.75	0.367725148307756\\
5.25	0.367725148307756\\
5.25	0.158082177430966\\
4.75	0.158082177430966\\
};
\addplot [color=red, line width=0.25mm, forget plot]
  table[row sep=crcr]{%
0.75	0.302361449931669\\
1.25	0.302361449931669\\
};
\addplot [color=red, line width=0.25mm, forget plot]
  table[row sep=crcr]{%
1.75	0.302815366364159\\
2.25	0.302815366364159\\
};
\addplot [color=red, line width=0.25mm, forget plot]
  table[row sep=crcr]{%
2.75	0.296891653595158\\
3.25	0.296891653595158\\
};
\addplot [color=red, line width=0.25mm, forget plot]
  table[row sep=crcr]{%
3.75	0.221235081465814\\
4.25	0.221235081465814\\
};
\addplot [color=red, line width=0.25mm, forget plot]
  table[row sep=crcr]{%
4.75	0.272209703872885\\
5.25	0.272209703872885\\
};
\end{axis}
\end{tikzpicture}%
}} &
  \subfloat[]{
    \resizebox{0.33\textwidth}{!}{
%
%
\begin{tikzpicture}
\definecolor{blue}{RGB}{76,100,135}
\definecolor{red}{RGB}{153,0,0}

\begin{axis}[%
width=4.634in,
height=3.612in,
at={(0.777in,0.487in)},
unbounded coords=jump,
xmin=0.5,
xmax=5.5,
xtick={1,2,3,4,5},
xticklabels={s-IK,s-IK2,m-IK,IK,e-IK},
ymin=0.0000012042434369705,
ymax=0.3,
ytick={ 0.1, 0.2},
yticklabels={0.1, 0.2},
ylabel style={font=\color{white!15!black}},
ylabel={$\sigma{}_{\text{min}}$},
ylabel near ticks,
ymajorgrids,
scale=0.5,
axis background/.style={fill=white},
title style={font=\bfseries},
title={Kinova Jaco},
legend style={legend cell align=left, align=left, draw=white!15!black}
]
\addplot [color=black, dashed, line width=0.25mm, forget plot]
  table[row sep=crcr]{%
1	0.243895778833598\\
1	0.252359338359006\\
};
\addplot [color=black, dashed, line width=0.25mm, forget plot]
  table[row sep=crcr]{%
2	0.237722115321223\\
2	0.245632388964833\\
};
\addplot [color=black, dashed, line width=0.25mm, forget plot]
  table[row sep=crcr]{%
3	0.226756353265445\\
3	0.253082402869304\\
};
\addplot [color=black, dashed, line width=0.25mm, forget plot]
  table[row sep=crcr]{%
4	0.204700024577003\\
4	0.249576596053442\\
};
\addplot [color=black, dashed, line width=0.25mm, forget plot]
  table[row sep=crcr]{%
5	0.206406493487979\\
5	0.251744998802753\\
};
\addplot [color=black, dashed, line width=0.25mm, forget plot]
  table[row sep=crcr]{%
1	0.141897381930051\\
1	0.198912453863033\\
};
\addplot [color=black, dashed, line width=0.25mm, forget plot]
  table[row sep=crcr]{%
2	0.128601069814952\\
2	0.192787964478507\\
};
\addplot [color=black, dashed, line width=0.25mm, forget plot]
  table[row sep=crcr]{%
3	0.0316290414726067\\
3	0.146190058079706\\
};
\addplot [color=black, dashed, line width=0.25mm, forget plot]
  table[row sep=crcr]{%
4	0.0125371060062565\\
4	0.11917171506304\\
};
\addplot [color=black, dashed, line width=0.25mm, forget plot]
  table[row sep=crcr]{%
5	0.0124719003755142\\
5	0.123231628170953\\
};
\addplot [color=black, line width=0.25mm, forget plot]
  table[row sep=crcr]{%
0.875	0.252359338359006\\
1.125	0.252359338359006\\
};
\addplot [color=black, line width=0.25mm, forget plot]
  table[row sep=crcr]{%
1.875	0.245632388964833\\
2.125	0.245632388964833\\
};
\addplot [color=black, line width=0.25mm, forget plot]
  table[row sep=crcr]{%
2.875	0.253082402869304\\
3.125	0.253082402869304\\
};
\addplot [color=black, line width=0.25mm, forget plot]
  table[row sep=crcr]{%
3.875	0.249576596053442\\
4.125	0.249576596053442\\
};
\addplot [color=black, line width=0.25mm, forget plot]
  table[row sep=crcr]{%
4.875	0.251744998802753\\
5.125	0.251744998802753\\
};
\addplot [color=black, line width=0.25mm, forget plot]
  table[row sep=crcr]{%
0.875	0.141897381930051\\
1.125	0.141897381930051\\
};
\addplot [color=black, line width=0.25mm, forget plot]
  table[row sep=crcr]{%
1.875	0.128601069814952\\
2.125	0.128601069814952\\
};
\addplot [color=black, line width=0.25mm, forget plot]
  table[row sep=crcr]{%
2.875	0.0316290414726067\\
3.125	0.0316290414726067\\
};
\addplot [color=black, line width=0.25mm, forget plot]
  table[row sep=crcr]{%
3.875	0.0125371060062565\\
4.125	0.0125371060062565\\
};
\addplot [color=black, line width=0.25mm, forget plot]
  table[row sep=crcr]{%
4.875	0.0124719003755142\\
5.125	0.0124719003755142\\
};
\addplot [color=blue, line width=0.25mm, forget plot]
  table[row sep=crcr]{%
0.75	0.198912453863033\\
0.75	0.243895778833598\\
1.25	0.243895778833598\\
1.25	0.198912453863033\\
0.75	0.198912453863033\\
};
\addplot [color=blue, line width=0.25mm, forget plot]
  table[row sep=crcr]{%
1.75	0.192787964478507\\
1.75	0.237722115321223\\
2.25	0.237722115321223\\
2.25	0.192787964478507\\
1.75	0.192787964478507\\
};
\addplot [color=blue, line width=0.25mm, forget plot]
  table[row sep=crcr]{%
2.75	0.146190058079706\\
2.75	0.226756353265445\\
3.25	0.226756353265445\\
3.25	0.146190058079706\\
2.75	0.146190058079706\\
};
\addplot [color=blue, line width=0.25mm, forget plot]
  table[row sep=crcr]{%
3.75	0.11917171506304\\
3.75	0.204700024577003\\
4.25	0.204700024577003\\
4.25	0.11917171506304\\
3.75	0.11917171506304\\
};
\addplot [color=blue, line width=0.25mm, forget plot]
  table[row sep=crcr]{%
4.75	0.123231628170953\\
4.75	0.206406493487979\\
5.25	0.206406493487979\\
5.25	0.123231628170953\\
4.75	0.123231628170953\\
};
\addplot [color=red, line width=0.25mm, forget plot]
  table[row sep=crcr]{%
0.75	0.228448694158491\\
1.25	0.228448694158491\\
};
\addplot [color=red, line width=0.25mm, forget plot]
  table[row sep=crcr]{%
1.75	0.217087002746571\\
2.25	0.217087002746571\\
};
\addplot [color=red, line width=0.25mm, forget plot]
  table[row sep=crcr]{%
2.75	0.195599669745134\\
3.25	0.195599669745134\\
};
\addplot [color=red, line width=0.25mm, forget plot]
  table[row sep=crcr]{%
3.75	0.161965055509587\\
4.25	0.161965055509587\\
};
\addplot [color=red, line width=0.25mm, forget plot]
  table[row sep=crcr]{%
4.75	0.165786183582426\\
5.25	0.165786183582426\\
};
\end{axis}
\end{tikzpicture}%
}} &
  \subfloat[]{
    \resizebox{0.33\textwidth}{!}{
%
%
\begin{tikzpicture}
\definecolor{blue}{RGB}{76,100,135}
\definecolor{red}{RGB}{153,0,0}

\begin{axis}[%
width=4.634in,
height=3.612in,
at={(0.777in,0.487in)},
unbounded coords=jump,
xmin=0.5,
xmax=5.5,
xtick={1,2,3,4,5},
xticklabels={s-IK,s-IK2,m-IK,IK,e-IK},
ymin=0.0000012042434369705,
ymax=0.345,
ytick={ 0.1, 0.2},
yticklabels={0.1, 0.2},
ylabel style={font=\color{white!15!black}},
ylabel={$\sigma{}_{\text{min}}$},
ylabel near ticks,
ymajorgrids,
scale=0.5,
axis background/.style={fill=white},
title style={font=\bfseries},
title={KUKA IIWA},
legend style={legend cell align=left, align=left, draw=white!15!black}
]
\addplot [color=black, dashed, line width=0.25mm, forget plot]
  table[row sep=crcr]{%
1	0.313520122659544\\
1	0.32931741504043\\
};
\addplot [color=black, dashed, line width=0.25mm, forget plot]
  table[row sep=crcr]{%
2	0.304261817363209\\
2	0.314948576585533\\
};
\addplot [color=black, dashed, line width=0.25mm, forget plot]
  table[row sep=crcr]{%
3	0.287438989113098\\
3	0.319991941281717\\
};
\addplot [color=black, dashed, line width=0.25mm, forget plot]
  table[row sep=crcr]{%
4	0.263891863943136\\
4	0.321216068565151\\
};
\addplot [color=black, dashed, line width=0.25mm, forget plot]
  table[row sep=crcr]{%
5	0.277993674185932\\
5	0.323275787927994\\
};
\addplot [color=black, dashed, line width=0.25mm, forget plot]
  table[row sep=crcr]{%
1	0.155103121739391\\
1	0.247192108951477\\
};
\addplot [color=black, dashed, line width=0.25mm, forget plot]
  table[row sep=crcr]{%
2	0.146657763405186\\
2	0.238250263874155\\
};
\addplot [color=black, dashed, line width=0.25mm, forget plot]
  table[row sep=crcr]{%
3	0.110509164791991\\
3	0.211184640575286\\
};
\addplot [color=black, dashed, line width=0.25mm, forget plot]
  table[row sep=crcr]{%
4	0.0187225457457219\\
4	0.16329875460117\\
};
\addplot [color=black, dashed, line width=0.25mm, forget plot]
  table[row sep=crcr]{%
5	0.0609630522322459\\
5	0.190341646249695\\
};
\addplot [color=black, line width=0.25mm, forget plot]
  table[row sep=crcr]{%
0.875	0.32931741504043\\
1.125	0.32931741504043\\
};
\addplot [color=black, line width=0.25mm, forget plot]
  table[row sep=crcr]{%
1.875	0.314948576585533\\
2.125	0.314948576585533\\
};
\addplot [color=black, line width=0.25mm, forget plot]
  table[row sep=crcr]{%
2.875	0.319991941281717\\
3.125	0.319991941281717\\
};
\addplot [color=black, line width=0.25mm, forget plot]
  table[row sep=crcr]{%
3.875	0.321216068565151\\
4.125	0.321216068565151\\
};
\addplot [color=black, line width=0.25mm, forget plot]
  table[row sep=crcr]{%
4.875	0.323275787927994\\
5.125	0.323275787927994\\
};
\addplot [color=black, line width=0.25mm, forget plot]
  table[row sep=crcr]{%
0.875	0.155103121739391\\
1.125	0.155103121739391\\
};
\addplot [color=black, line width=0.25mm, forget plot]
  table[row sep=crcr]{%
1.875	0.146657763405186\\
2.125	0.146657763405186\\
};
\addplot [color=black, line width=0.25mm, forget plot]
  table[row sep=crcr]{%
2.875	0.110509164791991\\
3.125	0.110509164791991\\
};
\addplot [color=black, line width=0.25mm, forget plot]
  table[row sep=crcr]{%
3.875	0.0187225457457219\\
4.125	0.0187225457457219\\
};
\addplot [color=black, line width=0.25mm, forget plot]
  table[row sep=crcr]{%
4.875	0.0609630522322459\\
5.125	0.0609630522322459\\
};
\addplot [color=blue, line width=0.25mm, forget plot]
  table[row sep=crcr]{%
0.75	0.247192108951477\\
0.75	0.313520122659544\\
1.25	0.313520122659544\\
1.25	0.247192108951477\\
0.75	0.247192108951477\\
};
\addplot [color=blue, line width=0.25mm, forget plot]
  table[row sep=crcr]{%
1.75	0.238250263874155\\
1.75	0.304261817363209\\
2.25	0.304261817363209\\
2.25	0.238250263874155\\
1.75	0.238250263874155\\
};
\addplot [color=blue, line width=0.25mm, forget plot]
  table[row sep=crcr]{%
2.75	0.211184640575286\\
2.75	0.287438989113098\\
3.25	0.287438989113098\\
3.25	0.211184640575286\\
2.75	0.211184640575286\\
};
\addplot [color=blue, line width=0.25mm, forget plot]
  table[row sep=crcr]{%
3.75	0.16329875460117\\
3.75	0.263891863943136\\
4.25	0.263891863943136\\
4.25	0.16329875460117\\
3.75	0.16329875460117\\
};
\addplot [color=blue, line width=0.25mm, forget plot]
  table[row sep=crcr]{%
4.75	0.190341646249695\\
4.75	0.277993674185932\\
5.25	0.277993674185932\\
5.25	0.190341646249695\\
4.75	0.190341646249695\\
};
\addplot [color=red, line width=0.25mm, forget plot]
  table[row sep=crcr]{%
0.75	0.291258774744165\\
1.25	0.291258774744165\\
};
\addplot [color=red, line width=0.25mm, forget plot]
  table[row sep=crcr]{%
1.75	0.276762167055828\\
2.25	0.276762167055828\\
};
\addplot [color=red, line width=0.25mm, forget plot]
  table[row sep=crcr]{%
2.75	0.272032636332949\\
3.25	0.272032636332949\\
};
\addplot [color=red, line width=0.25mm, forget plot]
  table[row sep=crcr]{%
3.75	0.21801940564599\\
4.25	0.21801940564599\\
};
\addplot [color=red, line width=0.25mm, forget plot]
  table[row sep=crcr]{%
4.75	0.24788234481664\\
5.25	0.24788234481664\\
};
\end{axis}

\end{tikzpicture}%
}} \\[-4ex]
  \subfloat[]{
    \resizebox{0.33\textwidth}{!}{
%
%
\begin{tikzpicture}
\definecolor{blue}{RGB}{76,100,135}
\definecolor{red}{RGB}{153,0,0}
\begin{axis}[%
width=4.634in,
height=3.612in,
at={(0.777in,0.487in)},
unbounded coords=jump,
xmin=0.5,
xmax=5.5,
xtick={1,2,3,4,5},
xticklabels={s-IK,s-IK2,m-IK,IK,e-IK},
ymin=0.4,
ymax=1.6,
ytick={ 0.95, 1.4},
yticklabels={0.7, 1.4},
ylabel style={font=\color{white!15!black}},
ylabel={$\sigma{}_{\text{max}}$},
ylabel near ticks,
ymajorgrids,
scale=0.5,
axis background/.style={fill=white},
title style={font=\bfseries},
legend style={legend cell align=left, align=left, draw=white!15!black}
]
\addplot [color=black, dashed, line width=0.25mm, forget plot]
  table[row sep=crcr]{%
1	1.21922094119238\\
1	1.52147102330177\\
};
\addplot [color=black, dashed, line width=0.25mm, forget plot]
  table[row sep=crcr]{%
2	1.2224352141838\\
2	1.52168048555983\\
};
\addplot [color=black, dashed, line width=0.25mm, forget plot]
  table[row sep=crcr]{%
3	1.22461740145553\\
3	1.52168707135491\\
};
\addplot [color=black, dashed, line width=0.25mm, forget plot]
  table[row sep=crcr]{%
4	1.16447392554441\\
4	1.52198430410788\\
};
\addplot [color=black, dashed, line width=0.25mm, forget plot]
  table[row sep=crcr]{%
5	1.21409296688066\\
5	1.52269895470725\\
};
\addplot [color=black, dashed, line width=0.25mm, forget plot]
  table[row sep=crcr]{%
1	0.556975116431575\\
1	0.604701181917684\\
};
\addplot [color=black, dashed, line width=0.25mm, forget plot]
  table[row sep=crcr]{%
2	0.562082587194757\\
2	0.702503273760661\\
};
\addplot [color=black, dashed, line width=0.25mm, forget plot]
  table[row sep=crcr]{%
3	0.507871027358999\\
3	0.710581770096504\\
};
\addplot [color=black, dashed, line width=0.25mm, forget plot]
  table[row sep=crcr]{%
4	0.472356939472077\\
4	0.620227682465255\\
};
\addplot [color=black, dashed, line width=0.25mm, forget plot]
  table[row sep=crcr]{%
5	0.499412014162291\\
5	0.71989835245019\\
};
\addplot [color=black, line width=0.25mm, forget plot]
  table[row sep=crcr]{%
0.875	1.52147102330177\\
1.125	1.52147102330177\\
};
\addplot [color=black, line width=0.25mm, forget plot]
  table[row sep=crcr]{%
1.875	1.52168048555983\\
2.125	1.52168048555983\\
};
\addplot [color=black, line width=0.25mm, forget plot]
  table[row sep=crcr]{%
2.875	1.52168707135491\\
3.125	1.52168707135491\\
};
\addplot [color=black, line width=0.25mm, forget plot]
  table[row sep=crcr]{%
3.875	1.52198430410788\\
4.125	1.52198430410788\\
};
\addplot [color=black, line width=0.25mm, forget plot]
  table[row sep=crcr]{%
4.875	1.52269895470725\\
5.125	1.52269895470725\\
};
\addplot [color=black, line width=0.25mm, forget plot]
  table[row sep=crcr]{%
0.875	0.556975116431575\\
1.125	0.556975116431575\\
};
\addplot [color=black, line width=0.25mm, forget plot]
  table[row sep=crcr]{%
1.875	0.562082587194757\\
2.125	0.562082587194757\\
};
\addplot [color=black, line width=0.25mm, forget plot]
  table[row sep=crcr]{%
2.875	0.507871027358999\\
3.125	0.507871027358999\\
};
\addplot [color=black, line width=0.25mm, forget plot]
  table[row sep=crcr]{%
3.875	0.472356939472077\\
4.125	0.472356939472077\\
};
\addplot [color=black, line width=0.25mm, forget plot]
  table[row sep=crcr]{%
4.875	0.499412014162291\\
5.125	0.499412014162291\\
};
\addplot [color=blue, line width=0.25mm, forget plot]
  table[row sep=crcr]{%
0.75	0.604701181917684\\
0.75	1.21922094119238\\
1.25	1.21922094119238\\
1.25	0.604701181917684\\
0.75	0.604701181917684\\
};
\addplot [color=blue, line width=0.25mm, forget plot]
  table[row sep=crcr]{%
1.75	0.702503273760661\\
1.75	1.2224352141838\\
2.25	1.2224352141838\\
2.25	0.702503273760661\\
1.75	0.702503273760661\\
};
\addplot [color=blue, line width=0.25mm, forget plot]
  table[row sep=crcr]{%
2.75	0.710581770096504\\
2.75	1.22461740145553\\
3.25	1.22461740145553\\
3.25	0.710581770096504\\
2.75	0.710581770096504\\
};
\addplot [color=blue, line width=0.25mm, forget plot]
  table[row sep=crcr]{%
3.75	0.620227682465255\\
3.75	1.16447392554441\\
4.25	1.16447392554441\\
4.25	0.620227682465255\\
3.75	0.620227682465255\\
};
\addplot [color=blue, line width=0.25mm, forget plot]
  table[row sep=crcr]{%
4.75	0.71989835245019\\
4.75	1.21409296688066\\
5.25	1.21409296688066\\
5.25	0.71989835245019\\
4.75	0.71989835245019\\
};
\addplot [color=red, line width=0.25mm, forget plot]
  table[row sep=crcr]{%
0.75	0.879143948324812\\
1.25	0.879143948324812\\
};
\addplot [color=red, line width=0.25mm, forget plot]
  table[row sep=crcr]{%
1.75	0.922278532446308\\
2.25	0.922278532446308\\
};
\addplot [color=red, line width=0.25mm, forget plot]
  table[row sep=crcr]{%
2.75	0.90648027402939\\
3.25	0.90648027402939\\
};
\addplot [color=red, line width=0.25mm, forget plot]
  table[row sep=crcr]{%
3.75	0.848076347538929\\
4.25	0.848076347538929\\
};
\addplot [color=red, line width=0.25mm, forget plot]
  table[row sep=crcr]{%
4.75	0.904839569637547\\
5.25	0.904839569637547\\
};
\end{axis}
\end{tikzpicture}%
}} &
  \subfloat[]{
    \resizebox{0.33\textwidth}{!}{
%
%
\begin{tikzpicture}
\definecolor{blue}{RGB}{76,100,135}
\definecolor{red}{RGB}{153,0,0}
\begin{axis}[%
width=4.634in,
height=3.612in,
at={(0.777in,0.487in)},
unbounded coords=jump,
xmin=0.5,
xmax=5.5,
xtick={1,2,3,4,5},
xticklabels={s-IK,s-IK2,m-IK,IK,e-IK},
ymin=0.00,
ymax=1.2,
ytick={ 0.5, 1},
yticklabels={0.5, 1},
ylabel style={font=\color{white!15!black}},
ylabel={$\sigma{}_{\text{max}}$},
ylabel near ticks,
ymajorgrids,
scale=0.5,
axis background/.style={fill=white},
title style={font=\bfseries},
legend style={legend cell align=left, align=left, draw=white!15!black}
]
\addplot [color=black, dashed, line width=0.25mm, forget plot]
  table[row sep=crcr]{%
1	0.776265131986956\\
1	1.04581117754036\\
};
\addplot [color=black, dashed, line width=0.25mm, forget plot]
  table[row sep=crcr]{%
2	0.773197031414908\\
2	1.04605692545888\\
};
\addplot [color=black, dashed, line width=0.25mm, forget plot]
  table[row sep=crcr]{%
3	0.775238130127563\\
3	1.04534074638063\\
};
\addplot [color=black, dashed, line width=0.25mm, forget plot]
  table[row sep=crcr]{%
4	0.781828246009887\\
4	1.04685544754586\\
};
\addplot [color=black, dashed, line width=0.25mm, forget plot]
  table[row sep=crcr]{%
5	0.794412440798894\\
5	1.04682534311602\\
};
\addplot [color=black, dashed, line width=0.25mm, forget plot]
  table[row sep=crcr]{%
1	0.208486888362497\\
1	0.493130176039249\\
};
\addplot [color=black, dashed, line width=0.25mm, forget plot]
  table[row sep=crcr]{%
2	0.21036219150917\\
2	0.533889587013207\\
};
\addplot [color=black, dashed, line width=0.25mm, forget plot]
  table[row sep=crcr]{%
3	0.218596003350196\\
3	0.473824207141795\\
};
\addplot [color=black, dashed, line width=0.25mm, forget plot]
  table[row sep=crcr]{%
4	0.219382975151073\\
4	0.443991089317249\\
};
\addplot [color=black, dashed, line width=0.25mm, forget plot]
  table[row sep=crcr]{%
5	0.227635173675067\\
5	0.461278409245692\\
};
\addplot [color=black, line width=0.25mm, forget plot]
  table[row sep=crcr]{%
0.875	1.04581117754036\\
1.125	1.04581117754036\\
};
\addplot [color=black, line width=0.25mm, forget plot]
  table[row sep=crcr]{%
1.875	1.04605692545888\\
2.125	1.04605692545888\\
};
\addplot [color=black, line width=0.25mm, forget plot]
  table[row sep=crcr]{%
2.875	1.04534074638063\\
3.125	1.04534074638063\\
};
\addplot [color=black, line width=0.25mm, forget plot]
  table[row sep=crcr]{%
3.875	1.04685544754586\\
4.125	1.04685544754586\\
};
\addplot [color=black, line width=0.25mm, forget plot]
  table[row sep=crcr]{%
4.875	1.04682534311602\\
5.125	1.04682534311602\\
};
\addplot [color=black, line width=0.25mm, forget plot]
  table[row sep=crcr]{%
0.875	0.208486888362497\\
1.125	0.208486888362497\\
};
\addplot [color=black, line width=0.25mm, forget plot]
  table[row sep=crcr]{%
1.875	0.21036219150917\\
2.125	0.21036219150917\\
};
\addplot [color=black, line width=0.25mm, forget plot]
  table[row sep=crcr]{%
2.875	0.218596003350196\\
3.125	0.218596003350196\\
};
\addplot [color=black, line width=0.25mm, forget plot]
  table[row sep=crcr]{%
3.875	0.219382975151073\\
4.125	0.219382975151073\\
};
\addplot [color=black, line width=0.25mm, forget plot]
  table[row sep=crcr]{%
4.875	0.227635173675067\\
5.125	0.227635173675067\\
};
\addplot [color=blue, line width=0.25mm, forget plot]
  table[row sep=crcr]{%
0.75	0.493130176039249\\
0.75	0.776265131986956\\
1.25	0.776265131986956\\
1.25	0.493130176039249\\
0.75	0.493130176039249\\
};
\addplot [color=blue, line width=0.25mm, forget plot]
  table[row sep=crcr]{%
1.75	0.533889587013207\\
1.75	0.773197031414908\\
2.25	0.773197031414908\\
2.25	0.533889587013207\\
1.75	0.533889587013207\\
};
\addplot [color=blue, line width=0.25mm, forget plot]
  table[row sep=crcr]{%
2.75	0.473824207141795\\
2.75	0.775238130127563\\
3.25	0.775238130127563\\
3.25	0.473824207141795\\
2.75	0.473824207141795\\
};
\addplot [color=blue, line width=0.25mm, forget plot]
  table[row sep=crcr]{%
3.75	0.443991089317249\\
3.75	0.781828246009887\\
4.25	0.781828246009887\\
4.25	0.443991089317249\\
3.75	0.443991089317249\\
};
\addplot [color=blue, line width=0.25mm, forget plot]
  table[row sep=crcr]{%
4.75	0.461278409245692\\
4.75	0.794412440798894\\
5.25	0.794412440798894\\
5.25	0.461278409245692\\
4.75	0.461278409245692\\
};
\addplot [color=red, line width=0.25mm, forget plot]
  table[row sep=crcr]{%
0.75	0.595945161057183\\
1.25	0.595945161057183\\
};
\addplot [color=red, line width=0.25mm, forget plot]
  table[row sep=crcr]{%
1.75	0.62621223210788\\
2.25	0.62621223210788\\
};
\addplot [color=red, line width=0.25mm, forget plot]
  table[row sep=crcr]{%
2.75	0.628898791494409\\
3.25	0.628898791494409\\
};
\addplot [color=red, line width=0.25mm, forget plot]
  table[row sep=crcr]{%
3.75	0.609216548871622\\
4.25	0.609216548871622\\
};
\addplot [color=red, line width=0.25mm, forget plot]
  table[row sep=crcr]{%
4.75	0.641416885531183\\
5.25	0.641416885531183\\
};
\end{axis}
\end{tikzpicture}%
}} &
  \subfloat[]{
    \resizebox{0.33\textwidth}{!}{
%
%
\begin{tikzpicture}
\definecolor{blue}{RGB}{76,100,135}
\definecolor{red}{RGB}{153,0,0}
\begin{axis}[%
width=4.634in,
height=3.612in,
at={(0.777in,0.487in)},
unbounded coords=jump,
xmin=0.5,
xmax=5.5,
xtick={1,2,3,4,5},
xticklabels={s-IK,s-IK2,m-IK,IK,e-IK},
ymin=0.0,
ymax=1.2,
ytick={ 0.5, 1},
yticklabels={0.5, 1},
ylabel style={font=\color{white!15!black}},
ylabel={$\sigma{}_{\text{max}}$},
ylabel near ticks,
ymajorgrids,
scale=0.5,
axis background/.style={fill=white},
title style={font=\bfseries},
legend style={legend cell align=left, align=left, draw=white!15!black}
]
\addplot [color=black, dashed, line width=0.25mm, forget plot]
  table[row sep=crcr]{%
1	0.892948629183011\\
1	1.13462139095611\\
};
\addplot [color=black, dashed, line width=0.25mm, forget plot]
  table[row sep=crcr]{%
2	0.904370248105116\\
2	1.13469216816606\\
};
\addplot [color=black, dashed, line width=0.25mm, forget plot]
  table[row sep=crcr]{%
3	0.902842588306145\\
3	1.13350419737216\\
};
\addplot [color=black, dashed, line width=0.25mm, forget plot]
  table[row sep=crcr]{%
4	0.876137470779122\\
4	1.1311051456753\\
};
\addplot [color=black, dashed, line width=0.25mm, forget plot]
  table[row sep=crcr]{%
5	0.905184519822138\\
5	1.12957986081918\\
};
\addplot [color=black, dashed, line width=0.25mm, forget plot]
  table[row sep=crcr]{%
1	0.399771805957401\\
1	0.495271761169949\\
};
\addplot [color=black, dashed, line width=0.25mm, forget plot]
  table[row sep=crcr]{%
2	0.42170244849072\\
2	0.542260599171831\\
};
\addplot [color=black, dashed, line width=0.25mm, forget plot]
  table[row sep=crcr]{%
3	0.287427220340739\\
3	0.550897121863626\\
};
\addplot [color=black, dashed, line width=0.25mm, forget plot]
  table[row sep=crcr]{%
4	0.285874098206592\\
4	0.528458842434518\\
};
\addplot [color=black, dashed, line width=0.25mm, forget plot]
  table[row sep=crcr]{%
5	0.287397152850082\\
5	0.57427180746504\\
};
\addplot [color=black, line width=0.25mm, forget plot]
  table[row sep=crcr]{%
0.875	1.13462139095611\\
1.125	1.13462139095611\\
};
\addplot [color=black, line width=0.25mm, forget plot]
  table[row sep=crcr]{%
1.875	1.13469216816606\\
2.125	1.13469216816606\\
};
\addplot [color=black, line width=0.25mm, forget plot]
  table[row sep=crcr]{%
2.875	1.13350419737216\\
3.125	1.13350419737216\\
};
\addplot [color=black, line width=0.25mm, forget plot]
  table[row sep=crcr]{%
3.875	1.1311051456753\\
4.125	1.1311051456753\\
};
\addplot [color=black, line width=0.25mm, forget plot]
  table[row sep=crcr]{%
4.875	1.12957986081918\\
5.125	1.12957986081918\\
};
\addplot [color=black, line width=0.25mm, forget plot]
  table[row sep=crcr]{%
0.875	0.399771805957401\\
1.125	0.399771805957401\\
};
\addplot [color=black, line width=0.25mm, forget plot]
  table[row sep=crcr]{%
1.875	0.42170244849072\\
2.125	0.42170244849072\\
};
\addplot [color=black, line width=0.25mm, forget plot]
  table[row sep=crcr]{%
2.875	0.287427220340739\\
3.125	0.287427220340739\\
};
\addplot [color=black, line width=0.25mm, forget plot]
  table[row sep=crcr]{%
3.875	0.285874098206592\\
4.125	0.285874098206592\\
};
\addplot [color=black, line width=0.25mm, forget plot]
  table[row sep=crcr]{%
4.875	0.287397152850082\\
5.125	0.287397152850082\\
};
\addplot [color=blue, line width=0.25mm, forget plot]
  table[row sep=crcr]{%
0.75	0.495271761169949\\
0.75	0.892948629183011\\
1.25	0.892948629183011\\
1.25	0.495271761169949\\
0.75	0.495271761169949\\
};
\addplot [color=blue, line width=0.25mm, forget plot]
  table[row sep=crcr]{%
1.75	0.542260599171831\\
1.75	0.904370248105116\\
2.25	0.904370248105116\\
2.25	0.542260599171831\\
1.75	0.542260599171831\\
};
\addplot [color=blue, line width=0.25mm, forget plot]
  table[row sep=crcr]{%
2.75	0.550897121863626\\
2.75	0.902842588306145\\
3.25	0.902842588306145\\
3.25	0.550897121863626\\
2.75	0.550897121863626\\
};
\addplot [color=blue, line width=0.25mm, forget plot]
  table[row sep=crcr]{%
3.75	0.528458842434518\\
3.75	0.876137470779122\\
4.25	0.876137470779122\\
4.25	0.528458842434518\\
3.75	0.528458842434518\\
};
\addplot [color=blue, line width=0.25mm, forget plot]
  table[row sep=crcr]{%
4.75	0.57427180746504\\
4.75	0.905184519822138\\
5.25	0.905184519822138\\
5.25	0.57427180746504\\
4.75	0.57427180746504\\
};
\addplot [color=red, line width=0.25mm, forget plot]
  table[row sep=crcr]{%
0.75	0.726971453866333\\
1.25	0.726971453866333\\
};
\addplot [color=red, line width=0.25mm, forget plot]
  table[row sep=crcr]{%
1.75	0.744314890067937\\
2.25	0.744314890067937\\
};
\addplot [color=red, line width=0.25mm, forget plot]
  table[row sep=crcr]{%
2.75	0.742622956018045\\
3.25	0.742622956018045\\
};
\addplot [color=red, line width=0.25mm, forget plot]
  table[row sep=crcr]{%
3.75	0.700098535361515\\
4.25	0.700098535361515\\
};
\addplot [color=red, line width=0.25mm, forget plot]
  table[row sep=crcr]{%
4.75	0.734560141183759\\
5.25	0.734560141183759\\
};
\end{axis}
\end{tikzpicture}%
}}
  \end{tabular}
  }
  \caption{Results of solving 200 random inverse kinematics (IK) problems; each column corresponds to IK solutions for a different common manipulator. The plots in the top row show the minimal singular values $\sigma_{\text{min}}$ of the manipulator Jacobian in the final configuration, while the plots in the bottom row show the maximal singular values $\sigma_{\text{max}}$. The two leftmost boxes in each plot, labeled s-IK and s-IK2, represent our method with for different choices of $\bSigma$. The box labeled m-IK corresponds to the method in \cite{dufour2017integrating}, while the box labeled IK shows the results without optimizing for singularity avoidance. Finally, the box labeled e-IK shows the results obtained when using a Euclidean metric.}\label{fig:exp2}
  \vspace*{-2mm}
\end{figure*}
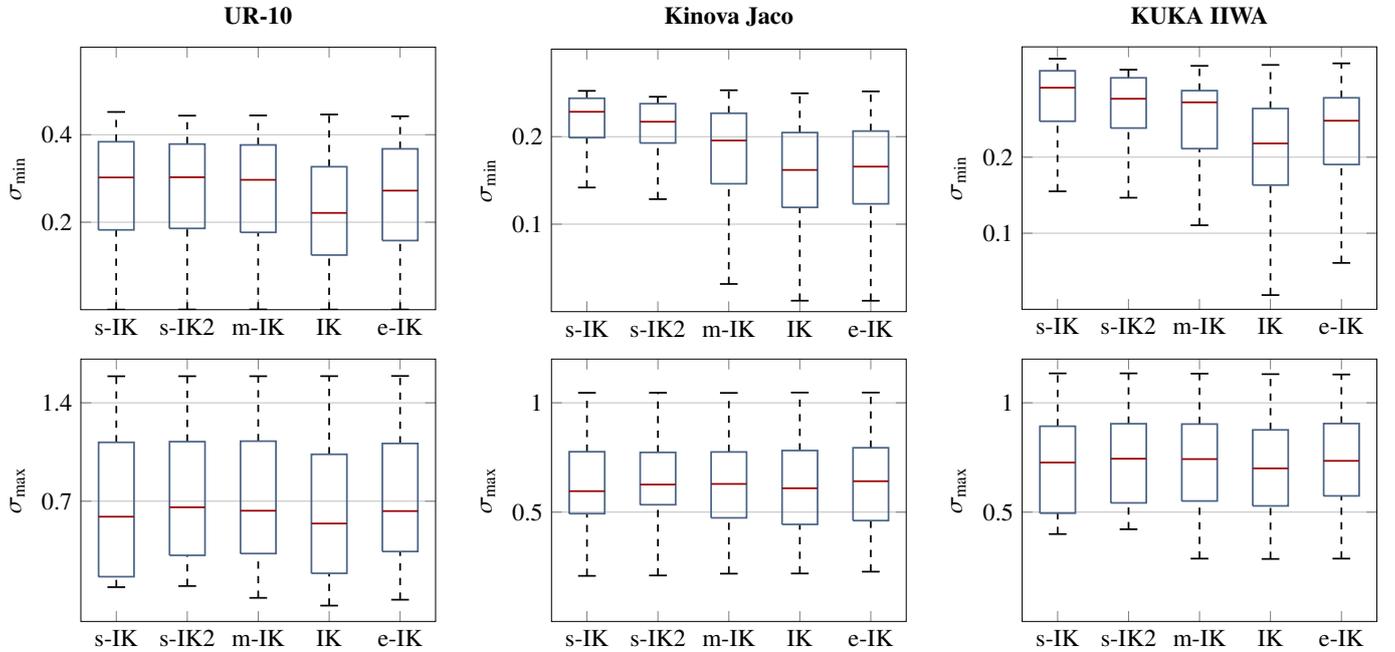
We begin by examining how the formulation given in \cref{eq:qp2} can be used to solve reaching tasks, where the end-effector needs to reach a desired goal position.
In our analysis, we consider the class of planar kinematic chains with an increasing number of DoF, as well as three robotic manipulators commonly used in collaborative, assistive, and research robotics.
We purposely avoid specifying a goal orientation in order to induce kinematic redundancy that can be used to optimize the singularity avoidance indices being tested.
The overall performance is determined by comparing the minimal and maximal singular values of the Jacobian in the final configuration, as these values provide a definitive indicator of singularity robustness for a given configuration.
We make the assumption that the joint limits and dynamic effects are accounted for by the velocity constraints at each iteration, making this problem similar to a standard inverse kinematics problem.
The experiment consists of performing 200 random (and randomly initialized) reaching tasks, while respecting upper and lower limits on joint velocities.

First, we examine the results for three, six, and nine DoF planar kinematic chains with joint velocities limited to $\frac{\pi}{8}\, \text{rad/s}$; the results are summarized by the box plots in \cref{fig:exp1}.
For methods s-IK, s-IK2, and m-IK, we chose $\alpha = 1$, since it produced good singularity avoidance results with a similar number of successes across the board.
The gradient of \cref{eq:euclidean}, used in the e-IK formulation, generally has a larger magnitude and so we use $\alpha=0.1$ to ensure numerical stability.
Examining the top row of \cref{fig:exp1}, we see that the method labeled s-IK, corresponding to the reference ellipsoid $\bSigma = k\mathbf{I}$ with $k = \mbox{Tr}(\mathbf{M}) \geq \sigma_{\text{max}}$ (updated at each iteration), achieves the highest median minimal singular value $\sigma_{\text{min}}$.
Moreover, increasing the number of DoF further amplifies this effect, as there is a larger space of solutions that can be explored due to the higher degree of redundancy.
This result is highly desirable from the perspective of singularity avoidance, since we are trying to avoid situations where the Jacobian is non-invertible.
However, minimizing $\xi$ in this case also results in $\mathbf{M}$ adopting a more spherical shape.
This can be seen in the bottom row of \cref{fig:exp1}, where s-IK produces a lower median maximal singular value $\sigma_{\text{max}}$ than s-IK2 or m-IK, reflecting the spherical shape of the reference ellipsoid.
We posit that the spherical shape results in a more uniform mobility profile for the end-effector, while not affecting the proximity to singular configurations.
Alternatively, by choosing $\bSigma = k\mathbf{M}$ with $k = 2$ (s-IK2), we achieve an overall increase of both the minimal and maximal singular values, very similar to that of m-IK.
Intuitively, choosing $k=2$ means that, at every iteration, we attempt to reach an ellipsoid that is twice the size of the current ellipsoid.
The results can be interpreted by observing that the gradient in this case has the form of \cref{eq:distgrad2}, which is very similar to the manipulability gradient~\cite{maric2019fast} used in m-IK.
The m-IK method from~\cite{dufour2017integrating} maximizes $\mbox{det}(\mathbf{M})$, which translates to maximizing the overall volume of the manipulability ellipsoid.
While this method outperforms the baseline IK approach, the median minimal singular value obtained using this method is noticeably smaller than that of s-IK.
Finally, the e-IK method outperforms only the baseline in terms of the minimal singular value, as it appears to prioritize maximizing the largest singular value.

We have also performed the same experiment using common six and seven DoF robots: the Universal Robots UR10, the Kinova Jaco manipulator, and the KUKA IIWA 14.
All joint velocities were again limited to at most $\frac{\pi}{8}\, \text{rad/s}$ in either direction and the gain value was increased to $\alpha = 10$ for s-IK, s-IK2, e-IK, and m-IK.
Examining the results in \cref{fig:exp2}, we note that the overall singular values are lower than that of the planar case.
This is because the movement of these robots is more constrained in three dimensions than the movement of the planar mechanisms in two dimensions.
Moreover, the majority of the translational mobility in these robots is produced by the first three joints, further exacerbating this phenomenon.
The top row of \cref{fig:exp2} again shows that the s-IK method, using a spherical reference ellipsoid, produces superior results, with s-IK2 coming in as a close second.
In the bottom row, we see that the maximum singular values are similar in all scenarios for all robots.

\subsection{Circular Path Tracking}\label{sec:circ_path}

\begin{figure}[!t]
  \centering
  \subfloat[]{ \includegraphics[width=0.375\columnwidth]{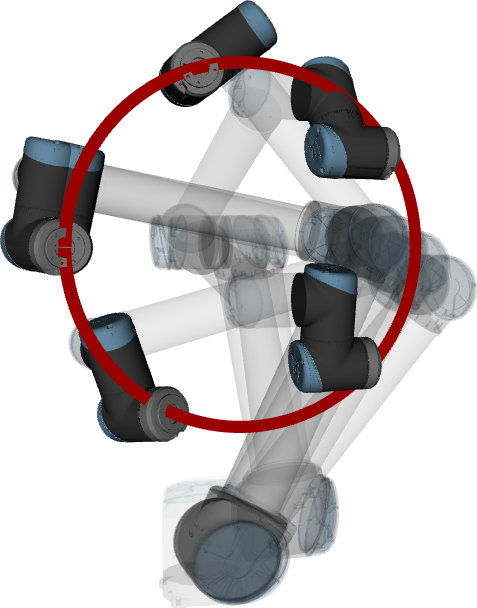}
  \label{fig:circle_arm}}\qquad\quad
  \subfloat[]{\includegraphics[width=0.325\columnwidth]{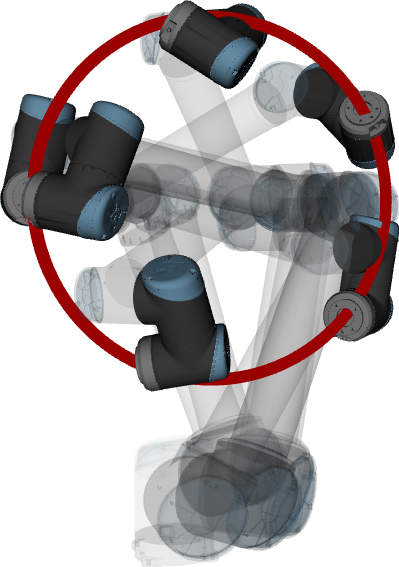}
  \label{fig:circle_arm_s}}
  \caption{(a) Following a circular path with the UR10 manipulator without attempting to avoid singular configurations. Note the large change in the wrist configuration when the manipulator reaches the top of the circle. (b) Following a circular path with the UR10 manipulator while using the singularity avoidance formulation s-IK. Note that the wrist and base configurations change more slowly, improving the conditioning of the Jacobian.}
\end{figure}

\begin{figure*}[!t]
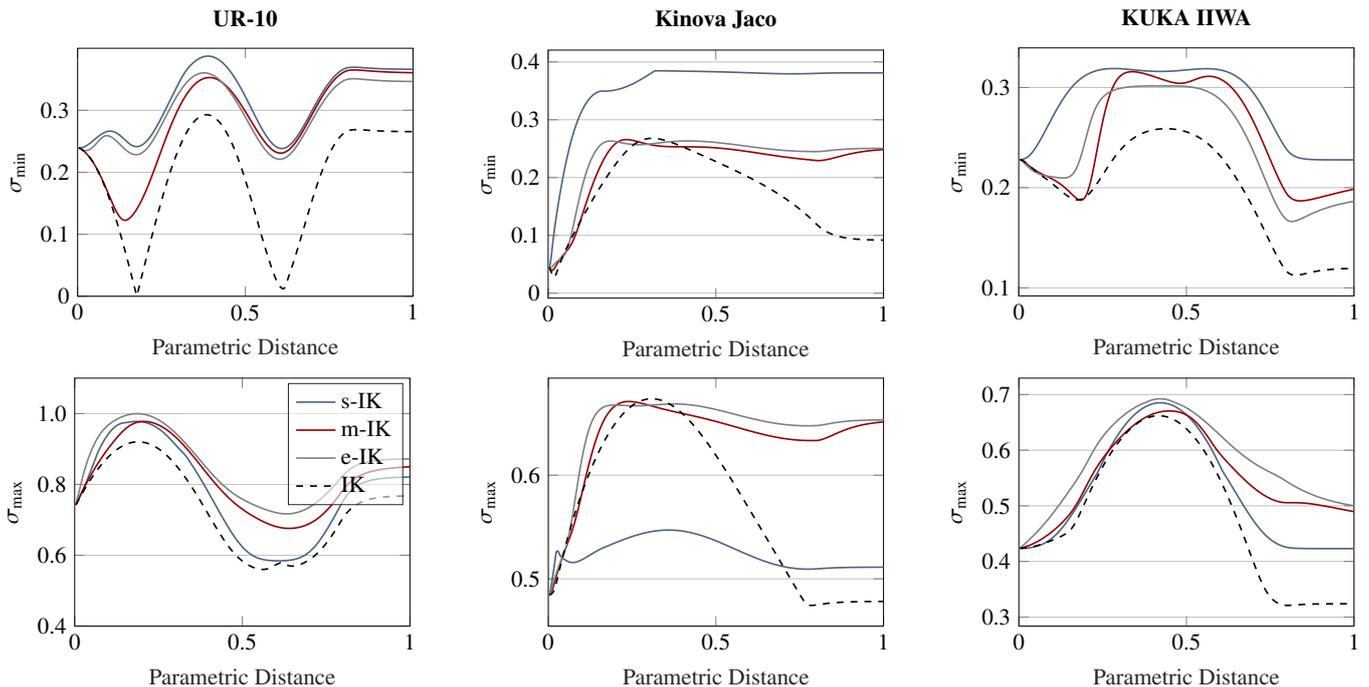

  \captionsetup[subfloat]{position=top,labelformat=empty}
  \centering
  \hspace*{-0.6cm}
  \resizebox{\textwidth}{!}{
  \begin{tabular}{@{}ccc@{}}
  \subfloat[]{
    \resizebox{0.7\columnwidth}{!}{\input{figs/sgv_min.tex}}}&
  \subfloat[]{
    \resizebox{0.7\columnwidth}{!}{\input{figs/sgv_min_jaco.tex}}} &
  \subfloat[]{
    \resizebox{0.7\columnwidth}{!}{\input{figs/sgv_min_kuka.tex}}} \\[-4ex]
  \subfloat[]{\resizebox{0.7\columnwidth}{!}{\input{figs/sgv_max.tex}}} &
  \subfloat[]{
    \resizebox{0.7\columnwidth}{!}{\input{figs/sgv_max_jaco.tex}}} &
  \subfloat[]{
    \resizebox{0.7\columnwidth}{!}{\input{figs/sgv_max_kuka.tex}}}
  \end{tabular}
  }
  \caption{Jacobian conditioning throughout the circular trajectory, parameterized by $t=\left[0,1\right]$. Minimal singular values $\sigma_{\text{min}}$ are displayed in the top row and maximal singular value $\sigma_{\text{max}}$ are displayed in the bottom row.}\label{fig:exp3}
\vspace*{-2mm}
\end{figure*}
In this experiment, we evaluated the performance of our singularity avoidance formulation in an operational space control scenario for several different manipulators by tracking a circular path with the end-effector.
All manipulators used in this experiment have six or more DoF, while the task required only the position of the end-effector to remain on the defined path at all times.
We are able to use the available kinematic redundancy to optimize the movement of each manipulator such that singular and near-singular configurations are avoided.
Again, the formulation in \cref{eq:qp2} is used to produce a locally optimal joint displacement at each iteration, and we compare the s-IK, m-IK, e-IK, and IK methods.
In \cref{fig:exp3} we see that the maximal and minimal singular values of the Jacobian vary throughout the trajectory (as executed by the chosen methods).
In the leftmost column, the regular IK method produces two nearly-singular configurations for the UR10, whereas all other methods avoid these singularities.
The singularities are indicated by two dips in $\sigma_{\text{min}}$ for the IK method, corresponding to the top and bottom of the circular path shown in \cref{fig:circle_arm}.
At these points, the wrist configuration of the manipulator in \cref{fig:circle_arm} shifts significantly in order to continue to follow the position reference---this is a clear indicator that the arm is passing near a singularity.
The m-IK method produces a trajectory closer to the first singularity than for s-IK or IK, since the m-IK maximizes the overall manipulability ellipsoid volume by prioritizing the increase of the two larger singular values.
Our method outperforms both the IK and m-IK methods by maintaining a $\sigma_{\text{min}}$ that is approximately two times larger than that produced by m-IK.
The trade-off can be seen by observing the bottom row, where it is clear that the $\sigma_{\text{max}}$ achieved by s-IK is somewhat lower than by m-IK or e-IK, while still outperforming IK.
The e-IK method performs surprisingly very well in this scenario, maintaining a $\sigma_{\text{min}}$ that is only slightly lower than that produced by s-IK, while achieving higher $\sigma_{\text{max}}$.
\cref{fig:circle_arm_s} shows the trajectories generated by s-IK; note the smaller variations in wrist movement, which suggests a better joint-to-operational space mapping in terms of end-effector position change compared to \cref{fig:circle_arm}.
Results for the same task performed by the Kinova Jaco arm are shown in the middle column of \cref{fig:exp3} and offer a contrasting example.
The performance of e-IK no longer matches s-IK in terms of singularity avoidance, as our method maintains a significantly higher $\sigma_{\text{min}}$ throughout the trajectory.
Interestingly, $\sigma_{\text{min}}$ and $\sigma_{\text{max}}$ reach a similar value for s-IK, which corresponds to the spherical shape of the reference ellipsoid used by this method.
The rightmost column of \cref{fig:exp3} shows that results for the KUKA-IIWA confirm the observed trend.
For all the robots, the baseline IK method is outperformed by all other methods, demonstrating that even using a `geometrically improper' criterion based on the Euclidean metric helps to avoid singularities.
Surprisingly, the performance of m-IK and e-IK is similar across all examples.

\begin{figure}[!t]
\centering
\includegraphics[trim=100 110 100 100,clip,width=0.7\columnwidth]{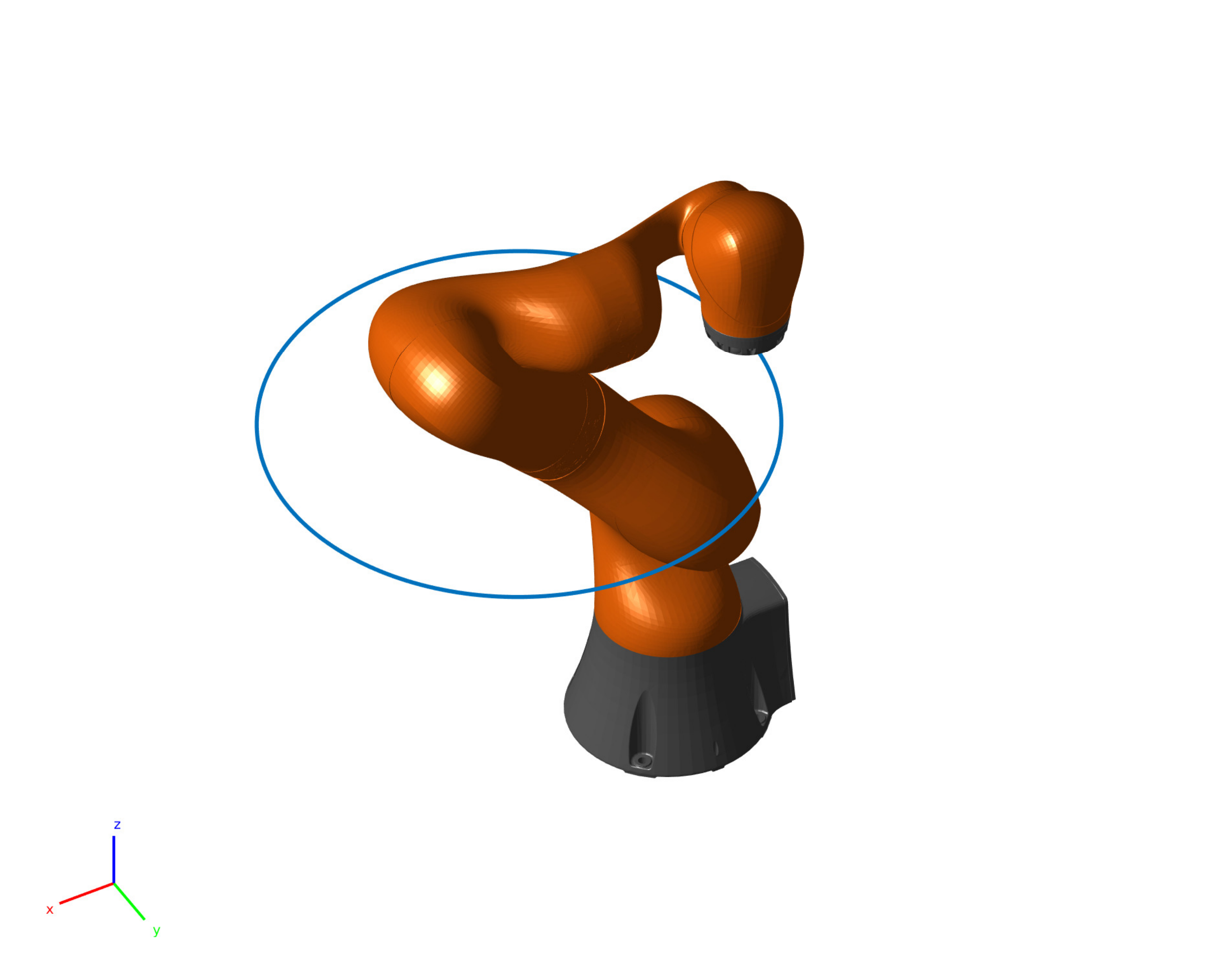}
\caption{Visualization of the circular path followed by the KUKA IIWA 14 manipulator. The end effector orientation is constrained to remain constant throughout the task.}
\label{fig:kuka_circle}
\vspace*{-2mm}
\end{figure}

\begin{figure*}[!t]
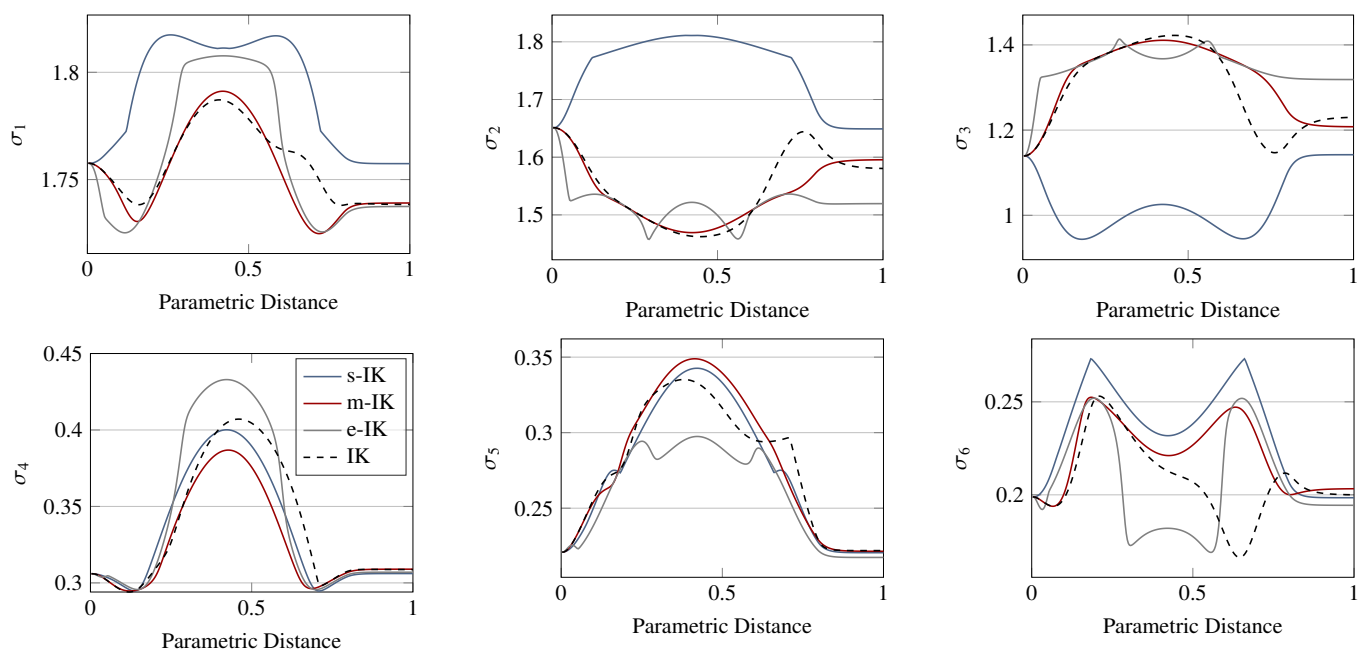

  \captionsetup[subfloat]{position=top,labelformat=empty}
  \centering
  \hspace*{-0.6cm}
  \resizebox{\textwidth}{!}{
  \begin{tabular}{@{}ccc@{}}
  \subfloat[]{\resizebox{0.7\columnwidth}{!}{\input{figs/experiment_2_new/svg_1.tex}}} &
  \subfloat[]{
    \resizebox{0.7\columnwidth}{!}{\input{figs/experiment_2_new/svg_2.tex}}} &
  \subfloat[]{
    \resizebox{0.7\columnwidth}{!}{\input{figs/experiment_2_new/svg_3.tex}}} \\[-4ex]
  \subfloat[]{
    \resizebox{0.7\columnwidth}{!}{\input{figs/experiment_2_new/svg_4.tex}}}&
  \subfloat[]{
    \resizebox{0.7\columnwidth}{!}{\input{figs/experiment_2_new/svg_5.tex}}} &
  \subfloat[]{
    \resizebox{0.7\columnwidth}{!}{\input{figs/experiment_2_new/svg_6.tex}}}
  \end{tabular}
  }
  \caption{Jacobian conditioning throughout the execution of a  circular trajectory with constant end-effector orientation for the KUKA IIWA 14 robot. The trajectory is  parameterized by $t=\left[0,1\right]$. Plots show how the Jacobian singular values $\sigma_{i}$ change as the robot follows the path. Singular values are indexed as follows" $\sigma_{1}\geq\sigma_{2}\geq\ldots\geq\sigma_{6}$.}\label{fig:exp4}
\vspace*{-4mm}
\end{figure*}

Finally, we have evaluated how all four methods perform when the path-following task also requires the end-effector orientation to remain fixed.
We perform this task with the seven DoF KUKA IIWA 14 robot, since it has the redundancy needed to optimize for singularity avoidance when tracking the full end-effector pose reference.
A visualization of the task can be seen in \cref{fig:kuka_circle}, while the changes of all Jacobian singular values are shown in \cref{fig:exp4}.
Our results indicate that using the baseline IK method results in a significant drop in $\sigma_{\text{min}}$ around $t=0.6$, signaling that the robot is near a singularity.
This singularity is avoided by the s-IK and m-IK methods, which produce similar changes of the lowest singular value.
On the other hand, the performance of e-IK is arguably worse than that of the baseline IK method, with the manipulator configuration remaining nearly singular from from $t=0.3$ to $t=0.55$.
We see that s-IK also results in larger $\sigma_{\text{max}} = \sigma_{1}$ and $\sigma_{2}$ than all the other methods.

\section{Conclusion and Future Work}

In this paper, we described a novel method for singularity avoidance that uses a well-known Riemannian metric on the manifold of SPD matrices to formulate a computationally tractable optimization criterion based on geodesic length.
We proved that our geometry-aware singularity index can be differentiated by computing directional derivatives using an identity from computational matrix analysis.
Moreover, we showed that various and geometrically distinct criteria can be derived from this formulation by changing a single parameter (i.e., reference ellipsoid) and that some choices result in robustness to failure modes that are common for  other indices.
We demonstrated that the proposed index can be integrated into a common optimization formulation of operational space reference tracking.
The experimental results indicate that our index consistently achieves the largest minimal singular values among all of the methods compared.
Moreover, we justified the use of the Riemannian metric by performing a comparison to an instance of our index that uses the standard Euclidean metric, which yields less consistent results and worse performance in terms of singularity avoidance.
Finally, it is important to note that there may be other choices of the reference ellipsoid suitable for singularity avoidance, as well as choices specifically tailored to a given task or kinematic structure.
We consider this one of the key advantages of our approach---our method can be tailored to a specific task to a greater degree than other indices such a the manipulability index.

As an avenue for future work, we note that the formulation presented herein can easily be integrated into existing control and planning pipelines and that it would be interesting to benchmark their performance.
We have previously integrated a manipulability maximization term into the trajectory optimization framework of \cite{maric2019fast}; we plan to integrate the proposed index into a similar formulation that is better suited for nonlinear objectives~\cite{petrovic2020cross}.
From a theoretical point of view, we are exploring possible equivalences between certain choices for the reference ellipsoid and existing singularity avoidance criteria.
Further, determining how a reference ellipsoid should be selected for a specific task may elucidate further advantages over more conventional methods.
%
\vspace*{-2mm}

\bibliographystyle{elsarticle-num}
\bibliography{thebibliography}

\end{document}